\documentclass[
	a4paper, 
	10pt, 
	unnumberedsections, 
	twoside, 
]{LTJournalArticle}
\usepackage{hyperref}
\usepackage{subcaption}
\usepackage{stfloats}
\usepackage{caption}

\usepackage{amsmath}
\title{Unified dimensionality reduction techniques in chronic liver disease detection} 

\author{%
	Anand Karna\textsuperscript{1}, Naina Khan\textsuperscript{1}, Rahul Rauniyar\textsuperscript{1}, Prashant Giridhar Shambharkar\textsuperscript{2}
}

\date{\footnotesize\textsuperscript{\textbf{1}}B. Tech Student, Computer Engineering Department, Delhi Technological University\\ \textsuperscript{\textbf{2}}Assistant Professor, Computer Engineering Department, Delhi Technological University\\
\textit{\small{*Corresponding Author: prashant.shambharkar@dtu.ac.in}}}

\renewcommand{\maketitlehookd}{%
	\begin{abstract}
		\noindent Globally, chronic liver disease continues to be a major health concern that requires precise predictive models for prompt detection and treatment. Using the Indian Liver Patient Dataset (ILPD) from the University of California at Irvine's UCI Machine Learning Repository, a number of machine learning algorithms are investigated in this study. The main focus of our research is this dataset, which includes the medical records of 583 patients, 416 of whom have been diagnosed with liver disease and 167 of whom have not. There are several aspects to this work, including feature extraction and dimensionality reduction methods like Linear Discriminant Analysis (LDA), Factor Analysis (FA), t-distributed Stochastic Neighbour Embedding (t-SNE), and Uniform Manifold Approximation and Projection (UMAP). The purpose of the study is to investigate how well these approaches work for converting high-dimensional datasets and improving prediction accuracy. To assess the prediction ability of the improved models, a number of classification methods were used, such as Multi-layer Perceptron, Random Forest, K-nearest neighbours, and Logistic Regression. Remarkably, the improved models performed admirably, with Random Forest having the highest accuracy of 98.31\% in 10-fold cross-validation and 95.79\% in train-test split evaluation. Findings offer important new perspectives on the choice and use of customized feature extraction and dimensionality reduction methods, which improve predictive models for patients with chronic liver disease.
	\end{abstract}
        \noindent \small{\textbf{Keywords:} Machine Learning, Preprocessing, Feature Extraction, LDA, t-SNE, Random Forest, Feature Integration, Liver Disease}
}

\begin{document}

\maketitle 


\section{1. Introduction}

\hspace{0.5cm}Over 1.7 million people globally pass away from a liver disease every year. The 11th most common cause of death and morbidity worldwide is cirrhosis (Cheemerla \& Balakrishnan, 2021). Prominent diseases of the liver include liver cirrhosis, hepatitis A, B, and C, fatty liver disease, and liver cancer. Over 75 million persons worldwide suffer from alcohol use disorders associated with liver disease out of the 2 billion drinkers (approx.) worldwide. In addition, 400 million adults have diabetes, 2 billion adults are overweight, and these conditions are risk factors for cancer and non-alcoholic fatty liver disease. Drug-induced liver problems are on the rise, and viral hepatitis is still very common (Asrani et al., 2019). Liver transplants, the second most common procedure, meet less than 10\% of global needs.

These are definitely fatal illnesses, but if caught early enough, they can be cured. The signs of liver disease usually appear only in the later stages of the illness so, early detection is difficult (Umbare et al., 2023). This is where machine learning techniques come in handy. They are widely employed in the healthcare industry, especially when it comes to the diagnosis and classification of certain diseases based on information about their features (Javaid et al., 2022). The increasing concern over chronic liver disease is what motivated this research. We employ cutting-edge methods to enhance prediction models on the Indian Liver Patient Dataset. In order to make timely treatment decisions for patients with chronic liver diseases, we are focusing on understanding how these techniques can improve diagnostic accuracy.
\begin{flushleft}
    \textbf{Contribution}
\end{flushleft}
\begin{itemize}
    \item The study introduces a novel and sophisticated methodology, incorporating IQR-based outlier replacement, oversampling, and a unique combination of dimensionality reduction techniques (LDA, FA, t-SNE, UMAP) followed by scaling.
    \item The study demonstrates outstanding results, particularly with the Random Forest algorithm, showcasing accuracy of 98.31\% and improved diagnostic precision.
    \item The study emphasizes the practical implications for clinical decision-making, providing nuanced insights for physicians to effectively apply sophisticated techniques in chronic liver disease diagnosis
    \item This research also contributes a valuable resource for practitioners and researchers, offering insights to enhance model performance in comparable healthcare settings.
\end{itemize}

The research paper's second section offers a thorough analysis of the body of existing literature as well as a thorough summary of all the pertinent studies. The investigation's methodology is explained in Section 3, where important details like dataset information, preprocessing methods, and feature extraction method integration are covered in detail. Section 4 presents the research outcomes and findings, providing an understanding of the conclusions drawn from the study process. The study's major findings and conclusions, which have been discussed throughout the paper, are summed up in the final section.
\section{2. Literature Survey}
A significant burden of chronic liver disease (CLD) exists worldwide; in 2017, there were 5.2 million cases, which resulted in 1.48 million deaths in 2019. The prevalence of cirrhosis has been rising since 1990. In 2019, disability-adjusted life-years came in at number 16 out of all diseases and at number 7 among those aged 50-74 (Liu \& Chen, 2022). A growing trend in alcohol consumption and cirrhosis linked to NAFLD contrasts (Younossi et al., 2016) with declining viral-associated burdens is seen. The use of predictive models to forecast the course of the disease, demonstrating the efficacy of ML in early identification. Additionally, effectiveness of machine learning in detecting problems has been shown, namely in differentiating between cirrhosis stages and forecasting the likelihood of hepatic decompensation (Popa et al., 2023). ML algorithms are essential for customizing treatment plans for individuals with CLD. Through the analysis of large datasets, specific patient features, and therapeutic strategy optimization based on expected outcomes, they offer individualized interventions. Patient results in the management of CLD have been greatly impacted by the use of machine learning, which has resulted in more focused and efficient treatments (Ahn et al., 2021).

When the high death rates associated with liver disorders were noticed, Bhupathi et al. (2022)  investigated machine learning models such as Support Vector Machines (SVM), Naïve Bayes, K-Nearest Neighbors (KNN), Linear Discriminant Analysis (LDA), and Classification and Regression Trees (CART). Notably, KNN achieved 91.7\% accuracy, while an autoencoder network demonstrated an even higher accuracy of 92.1\%. Dhayanand et al. (2015) found that when SVM (79.66\%) was applied to the Indian Liver Patient Records dataset, which contained 583 instances and 11 characteristics, it outperformed Naïve Bayes (61.28\%). Wu et al. (2019) highlighted in their contribution the features that are present in datasets for fatty liver disease (FLD), such as patient demographics including age, gender, blood pressure, glucose levels, and liver enzyme levels.  When data is collected using feature generation techniques in traditional machine learning systems, the input raw feature space is often highly dimensional and saturated with a large amount of useless feature information (Amin et al., 2023). The goal of linear discriminant analysis (LDA), a statistical technique for dimensionality reduction and classification, is to identify the linear feature combinations that most effectively divide data into distinct classes or groups (Batarseh \& Freeman, 2022). A statistical method called factor analysis (FA) is used to determine the latent components that underlie the correlations between observed data to comprehend the underlying structure among them (Tavakol \& Wetzel, 2020). A non-linear dimensionality reduction method called t-Distributed Stochastic Neighbour Embedding (t-SNE) is frequently used to visualize high-dimensional data in lower dimensions while maintaining local structures and emphasizing clusters or patterns in the data (Cai \& Ma, 2021). Another nonlinear dimensionality reduction method, noted for its capacity to successfully capture both local and global features in high-dimensional data, is called Uniform Manifold Approximation and Projection (UMAP). It is frequently used for the analysis and visualization of complex datasets.

Singh et al. (2020) performed a thorough assessment of classifier performance using ILPD and compared results with and without feature selection strategies. This study focused specifically on how feature selection affects classifier performance. Interestingly, when feature selection was removed, Logistic Regression showed the highest accuracy (72.50\%), but Random Forest outperformed with an accuracy of 74.36\% after feature selection was added. In order to clarify how feature selection strategies aid in the improvement of ILPD models, the evaluation comprised a variety of metrics, such as the number of examples that were successfully classified, the Kappa statistic, mean absolute error, and execution time. The study aimed to guide the development of more effective and efficient diagnostic models for liver health assessment by comparing classifiers with and without feature selection, thereby offering insights into the subtle effects of this approach. Wang et al. (2015) examined pulse signal analysis in cirrhosis patients, fatty liver disease (FLD) patients, and healthy individuals. Using sophisticated methods like harmonic fitting and learning algorithms, such as Principal Component Analysis (PCA), Least Squares (LS), and Least Absolute Shrinkage and Selection Operator (LASSO), the research sought to identify distinguishing patterns that would emphasize differences between the FLD/cirrhosis and healthy groups. With seven factors identified, this computer-aided diagnostic method showed great promise in the field of Traditional Chinese Medicine (TCM), with an accuracy rate of over 93\%. This highlights its potential to greatly improve clinical diagnostics and demonstrates its applicability as a useful instrument in the evaluation of disorders related to the liver.

Kumar and Sahoo (2013) used a dataset with 583 individuals to present a Rule-Based Classification Model (RBCM) that combines rule-based and data mining methods for liver disease prediction. Several ML, such as Support Vector Machines (SVM), Rule Induction (RI), Decision Trees (DT), Naïve Bayes (NB), and Artificial Neural Networks (ANN), were applied during their investigation. The Decision Tree (DT) demonstrated superior performance compared to the other methods, with remarkable results of 98.46\% accuracy, 95.7\% sensitivity, 95.28\% specificity, and a Kappa coefficient of 0.983. The study underscored the efficacy of the Rule-Based Classification Model, highlighting its superiority over non-rule-based models. The remarkable efficacy of DT highlighted the importance of applying rule-based methods when discussing the prognosis of liver disease. In addition to demonstrating the Rule-Based Classification Model's strong predictive abilities, this research established the model as a useful resource for medical decision-making regarding the prognosis of liver disease. Trigka et al. (2023) distinguished between "Liver-Disease" (LD) and "Non-Liver-Disease" (Non-LD) classes in their carefully constructed classification problem on liver disease prediction. A balanced dataset of 828 participants was produced by using the oversampling method (SMOTE) to address class imbalance. Afterwards, the significance of each feature was assessed using the three feature ranking techniques of Pearson Correlation, Gain Ratio, and Random Forest. Direct bilirubin (DB) was found to be the most significant feature after a Pearson correlation analysis showed significant relationships between a few features. In summary, the most crucial features—DB, TB, and SGOT—were used for the models' training and validation. With the use of multiple ML models and ensemble approaches, the developed predictive model showed impressive results in terms of accuracy (80.1\%), precision (80.4\%), recall (80.1\%), and area under the curve (AUC) (88.4\% post-SMOTE with 10-fold cross-validation).

Muthuselvan et al. (2018) investigated the use of machine learning algorithms like Naïve Bayes, K-Star, J48, and Random Tree for the classification of liver disease. The study concentrated on a dataset that included 11 liver health-related parameters and was obtained from the northeastern Indian state of Andhra Pradesh. In order to prepare the dataset for machine learning analyses, it underwent extensive exploration and pre-processing in WEKA. This included actions like handling missing values, data discretization, and converting numeric values to nominal ones. The unique methodology of the study consisted of analyzing various algorithms to identify the subtleties of their performances. The results showed that K-Star executed faster (less than 1 second) and had a reasonably competitive accuracy, while Random Tree had the highest accuracy (74.2\%) but required a longer execution time (0.05 secs). The study's focus on a variety of algorithms and real-world applicability, along with WEKA's user-friendly environment, provided insightful information for researchers examining machine learning within the framework of liver disease prediction as well as educators. 

Several classification algorithms, such as the Naive Bayesian Classifier (NBC), K-Nearest Neighbors (K-NN), and C4.5, were used in the study by Babu et al. (2016) to assess how well they predicted liver disease. These algorithms obtained classification accuracies of 0.56, 0.64, and 0.69 on the original ILPD dataset, respectively. But the researchers used a thorough approach that included feature selection and clustering, which resulted in the development of an updated dataset (NDS). The updated dataset showed significant gains in classification accuracy: 0.90, 0.95, and 0.93 for K-NN, C4.5, and NBC, respectively. The need to address issues with inconsistent data and low accuracy in clustering and classification techniques led to the selection of this specific methodology. The number of attributes or dimensions is a key factor that affects how well clustering algorithms perform, as the researchers discovered. They used feature selection techniques to lower dimension and then put the K-Means clustering algorithm into practice for unsupervised partitioning in order to lessen these difficulties. A refined training dataset was produced by validating the resulting clusters using cross-validation and expert intuition. The improved results on the NDS dataset highlight how well this methodology works to optimize liver disease prediction models. 

In a recent study, Sujith et al. (2023) used the ILPD dataset in a Jupyter Notebook environment to classify liver diseases using a variety of machine learning techniques, such as Convolutional Neural Networks (CNN), Recurrent Neural Networks (RNN), Artificial Neural Networks (ANN), and Logistic Regression. Among the methods investigated, CNN achieved the highest accuracy (67\%) and precision (71\%) when optimized with Adamax, making it the most successful model. The choice to employ Adamax is in line with its established efficacy in neural network model optimization, specifically in managing sparse gradients and fostering stability throughout the training process. The researchers' selection of Adamax is consistent with its broad use in machine learning applications and demonstrates its suitability for the particular difficulties associated with classifying liver disease in their study. In order to predict liver disease based on different attributes, Sivasangari et al. (2020) investigated the use of three different machine learning algorithms: SVM, DT and RF. The suggested approach called for gathering data from the UCI machine learning library, replacing missing values through preprocessing, and validating the model through a split train-test approach. The experimental results, which included confusion matrices and quantitative assessments of accuracy, precision, and recall, demonstrated the usefulness of the models. SVM achieved the highest accuracy (95.18\%) according to the comparative analysis, while RF showed perfect precision. The study emphasized the significance of algorithmic selection and evaluation metrics, and it provided insightful information about liver disease prediction through ML.
\section{3. Methodology}
\begin{figure}[h] 
    \centering
    \includegraphics[width=0.9\linewidth]{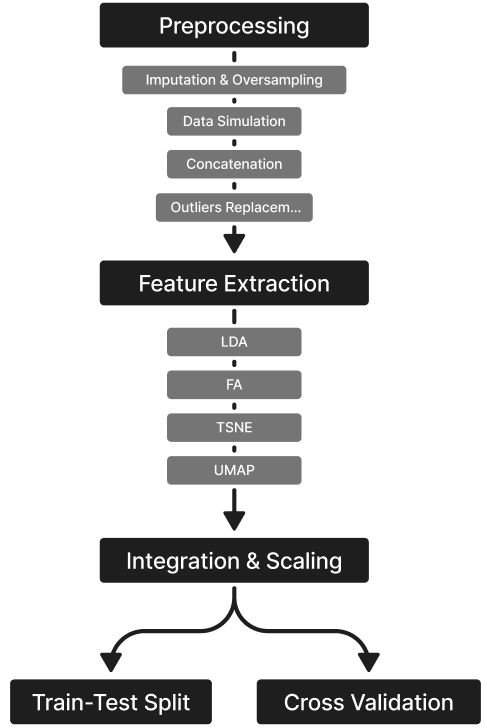}
    \caption{A flow diagram depicting the methodology used in the study}
    \label{fig:methodology}
\end{figure}

\hspace{0.5cm}In the course of this investigation, the Indian Liver Disease Dataset (ILPD) was subjected to a thorough preprocessing regimen, encompassing imputation and oversampling, data simulation, concatenation, and outlier replacement. Using dimensionality reduction methods, key patterns supporting the classification of liver disease were extracted. A strong protocol for efficient data handling was formed by this methodological approach, which also included operations like standard scaling, potential data integration, and outlier management. A rigorous train-test split and cross-validation approach was used during the evaluation phase to ensure the reliability of predictive models and reduce the biases associated with the investigation. This methodical and condensed approach made it easier to create a thorough framework for efficient data processing and model building. Figure \ref{fig:methodology} visually represents the research methodology.
\subsection{3.1. Dataset}
\hspace{0.5cm}A useful resource that is often obtained from the UCI Machine Learning Repository is the Indian Liver Patient Dataset (ILPD), which has patient medical attributes mostly pertaining to liver health. Many characteristics are usually included in this dataset, including age, gender, liver function tests, symptoms, alcohol intake, hepatitis history, blood test findings (bilirubin levels, liver enzymes), and symptoms. Its target variable shows whether a diagnosis of liver disease has been made (liver condition: 1) or not (liver condition: 2 or 0). The ILPD dataset, a moderately substantial collection of patient records, is used as a standard for machine learning research, facilitating the creation and assessment of models intended to predict the presence or absence of liver disease. We converted the 'Target' variable into a binary format, which typically indicates whether or not liver disease has been diagnosed.With this modification, the values 0 and 1 were mapped to represent the presence and absence of liver conditions, respectively.

\begin{table}[h]
\centering
\caption{Indian Liver Patient Dataset Feature Description}
\label{tab:variables}
{\footnotesize
\begin{tabular}{|p{0.9cm}|p{1.4cm}|p{2.8cm}|p{1cm}|}
\hline
\textbf{Variable Name} & \textbf{Type} & \textbf{Description} & \textbf{Missing Values} \\ \hline
Age                     & Integer        & Age                  & no                     \\ \hline
Gender                  & Categorical         & Gender               & no                     \\ \hline
TB                      & Real Number     & Total Bilirubin      & no                     \\ \hline
DB                      & Real Number     & Direct Bilirubin     & no                     \\ \hline
Alkphos                 & Integer        & Alkaline Phosphotase & no                     \\ \hline
Sgpt                    & Integer        & Alamine Aminotransferase & no                 \\ \hline
Sgot                    & Integer        & Aspartate Aminotransferase & no               \\ \hline
TP                      & Real Number     & Total Proteins       & no                     \\ \hline
ALB                     & Real Number     & Albumin              & no                     \\ \hline
A/G Ratio               & Real Number    & Albumin and Globulin Ratio & 4               \\ \hline
Target                  & Binary        & Absence or presence of disease & no               \\ \hline
\end{tabular}
}
\end{table}

\subsection{3.2. Preprocessing}
\begin{flushleft}
\textbf{3.2.1 Imputation and Oversampling}
\end{flushleft}
\hspace{0.5cm}We methodically preprocessed the data to reduce class imbalance within the dataset. This included removing unnecessary columns i.e., Gender as it is categorical, handling missing values, and uniformly converting data types. We used the mean of the 'ALB' column to impute the missing values in the 'A/G' column (McKnight et al., 2007). With RandomOverSampler, we were able to effectively balance the distribution of classes. After oversampling, the class distribution was 1334 (0) and 1330 (1). In order to reduce bias resulting from missing values and improve the accuracy of statistical measures, imputation was performed before outlier replacement. In order to mitigate biases in minority class representation, oversampling following imputation guaranteed that synthetic instances reflected a more comprehensive dataset.

\begin{flushleft}
\textbf{3.2.2. Data Simulation}
\end{flushleft}

We generated a synthetic dataset with 1000 samples and 10 features, where 5 features provided relevant information, 3 were redundant, and 2 were repeated. This dataset was produced specifically for a binary classification assignment. The synthetic dataset was then concatenated with the original dataset.

Let, the dataset matrix is represented by \(X\), where a feature is represented by a column and a sample by a row. This is represented mathematically as \(X \in \mathbb{R}^{1000 \times 10}\). \(X_{\text{info}}\) denotes the informative features, which are intended to hold pertinent data, while \(X_{\text{red}}\) stands for the redundant features. The formulation in can be written as follows:

\[X_{\text{info}} \in \mathbb{R}^{1000 \times 5}, \quad X_{\text{red}} \in \mathbb{R}^{1000 \times 3}\]

Additionally, \(X_{\text{rep}}\) represents the repeated features, which are duplicated to create variation in the dataset. Mathematically, this is represented as:

\[X_{\text{rep}} \in \mathbb{R}^{1000 \times 2}\]

This mathematical framework allowed for careful exploration of the behavior of machine learning models. Informative features aim to capture meaningful patterns, whereas redundant and repeated features pose challenges for models to process. Our understanding of how various types of information impact algorithms has improved as a result of this methodical approach, which has helped with model design and training methods. Furthermore, the learned lessons have useful ramifications for improving model applications in various real-world contexts.

\begin{flushleft}
\textbf{3.2.3. Concatenation}
\end{flushleft}

This operation was beneficial as it vertically stacked the samples from both datasets, creating a larger dataset that incorporated information from both sources. By utilizing the combined knowledge from the simulated and original datasets, it made it possible to conduct a more thorough analysis and train the model. 
\[ X_{\text{original}} \in \mathbb{R}^{n_{\text{original}} \times m} \]

\[ X_{\text{simulated}} \in \mathbb{R}^{n_{\text{simulated}} \times m} \]

\[ X_{\text{combined}} = \begin{bmatrix} X_{\text{original}} \\ X_{\text{simulated}} \end{bmatrix} \hspace{1.5cm}(1) \]
\[ X_{\text{combined}} \in \mathbb{R}^{(n_{\text{original}} + n_{\text{simulated}}) \times m} \]

\[ y_{\text{combined}} = \begin{bmatrix} y_{\text{original}} \\ y_{\text{simulated}} \end{bmatrix} \hspace{1.5cm}(2)\]
\[ y_{\text{combined}} \in \mathbb{R}^{n_{\text{original}} + n_{\text{simulated}}} \]

Aligning the target labels with the combined dataset was essential for accurate analysis and model training. The subsequent steps in the data processing pipeline were made easier by the mathematical process that guaranteed consistency in the dataset structure.

\begin{flushleft}
\textbf{3.2.4. Outliers Replacement}
\end{flushleft}

The Interquartile Range (IQR) is a robust statistical measure defined as the difference between the third quartile ($Q3$) and the first quartile ($Q1$):
\[
IQR = Q3 - Q1 \hspace{1.5cm}(3)
\]

The values known as quartiles split a dataset into four equal sections. The 25th percentile of the data is represented by the first quartile ($Q1$), and the 75th percentile by the third quartile ($Q3$). Because it is insensitive to extreme values or dataset outliers, the IQR is a reliable indicator of spread.
By identifying the distribution of the central 50\% of the data, it can withstand outliers. Lower and upper bounds are computed in order to identify outliers:
\[
\text{Lower Bound:}~  = Q1 - k \times IQR \hspace{1.5cm}(4)
\]
\[
\text{Upper Bound:}~ = Q3 + k \times IQR \hspace{1.5cm}(5)
\]
where $k$ (1.5 or 3) determines sensitivity.

To ensure reliability, outliers were then found and replaced with second quartile values using the IQR approach. At the same time, second quartile values were used to imputation for missing values. These methodical processing steps produced a refined dataset with IQR-based outlier detection, establishing the foundation for further analytical and modeling efforts.

\subsection{3.3. Feature Extraction}
\hspace{0.5cm}Feature extraction stands out as a crucial step in analysis as it means taking better, more significant characteristics out of the raw data in order to reduce dimension. This method seeks to improve data management and increase the effectiveness of machine learning models by eliminating unnecessary or superfluous features while keeping relevant information (Alpaydin, 2014). There are several approaches in the field of feature extraction, which include both linear and non-linear dimensionality reduction methods. When integrated, they provide a holistic method for improving data representation. Combining linear approaches for dimensionality reduction first and then using non-linear approaches enables a cooperative combination that captures both global and local data structures (Muixí et al., 2023). The combination of Linear Discriminant Analysis and Factor Analysis with non-linear methods like t-distributed Stochastic Neighbor Embedding and Uniform Manifold Approximation and Projection, which is the driving factor of the investigation, is reflected.

\begin{flushleft}
\textbf{3.3.1. Linear Discriminant Analysis}
\end{flushleft}

The goal of LDA is to maximize class separability while producing a lower-dimensional space (Tharwat et al., 2017). This is accomplished by determining the axes for classification jobs that retain the most amount of information. LDA prioritizes two important goals: increasing the between-class scatter, which guarantees that distinct class clusters are farther apart, and limiting the within-class scatter, which ensures that data points within the same class are closer together. The number of projected axes is determined by either the number of features or the number of unique classes reduced by one in order to prevent overfitting and capture the most discriminative information. LDA maximizes the performance of classification algorithms by preserving important class-specific information while identifying a transformation that best divides classes in the smaller feature space. This is done by computing the scatter matrices reflecting within-class and between-class variances.

The within-class scatter matrix is defined as:
\[ S_w = \sum_{i=1}^{K} \sum_{\mathbf{x} \in X_i} (\mathbf{x} - \boldsymbol{\mu}_i)(\mathbf{x} - \boldsymbol{\mu}_i)^T \hspace{1.5cm}(6)\]

The between-class scatter matrix is defined as:
\[ S_b = \sum_{i=1}^{K} N_i (\boldsymbol{\mu}_i - \boldsymbol{\mu})(\boldsymbol{\mu}_i - \boldsymbol{\mu})^T \hspace{1.5cm}(7) \]

Where:

- \(N\) = Total number of samples

- \(K\) = Number of classes

- \(\mathbf{x}_i\) = Samples from the \(i^{th}\) class

- \(\boldsymbol{\mu}_i\) = Mean of samples from the \(i^{th}\) class

- \(S_w\) = Within-class scatter matrix

- \(S_b\) = Between-class scatter matrix

- \(S_t = S_w + S_b\) (Total scatter matrix)

By resolving the generalized eigenvalue problem, LDA seeks to determine the projection vector \(\mathbf{w}\) that maximizes the ratio of between-class scatter to within-class scatter:
\[ S_w^{-1}S_b \cdot \mathbf{w} = \lambda \mathbf{w} \hspace{1.5cm}(8)\]
Where \(\mathbf{w}\) is the projection vector and \(\lambda\) is the eigenvalue associated with that vector. A lower-dimensional subspace that offers the greatest separation between distinct classes is created from the data using this projection vector, \(\mathbf{w}\).

\begin{flushleft}
    \textbf{3.3.2. Factor Analysis}
\end{flushleft}

By identifying latent variables—referred to as factors—that explain correlations between observed variables, FA is a statistical technique used to reduce the dimension of a dataset (Smelser \& Baltes, 2001). Its operation is predicated on the idea that fewer unobserved factors have an impact on observed variables. FA seeks to identify a lower-dimensional representation that captures most of the variability in the data and to summarize the underlying structure by estimating the associations between the observed variables using the covariance matrix.The number of factors to extract is pre-specified, in our case 3, in order to reveal the key elements of the dataset. Factor loadings, which show the degree of correlation between latent factors and observed variables, are calculated by FA. FA also estimates an error term, which is the unexplained variance or noise in the dataset that cannot be explained by the factors that have been found. This error term offers insights into the variability not captured by the extracted factors and aids in evaluating the model fit and understanding the residual variance in the dataset.

Using the covariance matrix, FA calculates the relationships between the observed variables in an effort to find a lower-dimensional representation that summarizes the underlying structure and captures the majority of the variability in the data (Johnson \& Wichern, 2007). The observed variables' covariance matrix (\(S\)) can be decomposed into two components representing common variance due to factors and unique variance:
\[ S = \mathbf{LL}^T + \boldsymbol{\Psi} \hspace{1.5cm}(9) \]
Where:

- \(S\) is the observed variables' covariance matrix.

- \(\mathbf{L}\) represents factor loadings.

- \(\boldsymbol{\Psi}\) signifies the diagonal matrix representing unique variances.

FA also estimates an error term, which is the unexplained variance or noise in the dataset that cannot be explained by the factors that have been found. This error term helps to clarify the residual variance in the dataset and evaluate the model fit. The estimation of the error term (\(\boldsymbol{\Psi}\)) is derived from the diagonal elements of the residual covariance matrix:
\[ \boldsymbol{\Psi} = \text{diag}(\boldsymbol{\Sigma} - \mathbf{LL}^T) \hspace{1.5cm}(10) \]
Where:

- \(\boldsymbol{\Sigma}\) is the sample covariance matrix of the observed variables.

- \(\mathbf{LL}^T\) represents the estimated common variance due to factors.

\begin{flushleft}
    \textbf{3.3.3. t-Distributed Stochastic Neighbour Embedding}
\end{flushleft}

We applied t-SNE, a dimensionality reduction technique, to combined features that were obtained from Linear Discriminant Analysis (LDA) and Factor Analysis (FA). To create a combined feature set, the reduced features that were extracted independently from LDA (X\_lda) and FA (X\_fa) were first concatenated  and then, t-SNE was then used. It was instantiated with three output dimensions and a fixed random state. t-SNE models high-dimensional data into a lower-dimensional space while maintaining the data points' local structure (Zhou et al., 2018). In order to facilitate visualization and investigation of underlying patterns or clusters within the data, it attempts to represent each high-dimensional data point in a lower-dimensional space, with dissimilar data points placed farther away and comparable data points closer together. As a result, the reduced features from the concatenated FA and LDA features obtained via t-SNE weres stored in the resulting X\_tsne.
The joint probability $ p_{ij} $ that a data point $ x_i $ would pick $ x_j $ as its neighbor in the high-dimensional space is calculated using a Gaussian distribution with the similarities  $ \text{sim}_i(x_j) $ (van der Maaten et al., 2008):

\[
p_{ij} = \frac{\exp(-\lVert x_i - x_j \rVert^2 / 2\sigma_i^2)}{\sum_{k \neq i} \exp(-\lVert x_i - x_k \rVert^2 / 2\sigma_i^2)} \hspace{1.5cm}(11)
\]

where $ \sigma_i $ is the Gaussian distribution's variance computed using the high-dimensional space's data point distances.

A Student's t-distribution with degrees of freedom $ \text{df} $ and the Euclidean distances $ \lVert y_i - y_j \rVert^2 $ between the low-dimensional representations are used to determine the joint probability $ q_{ij} $ that $ x_i $ would select $ x_j $ as its neighbor in the low-dimensional space:

\[
q_{ij} = \frac{(1 + \lVert y_i - y_j \rVert^2)^{-1}}{\sum_{k \neq i} (1 + \lVert y_i - y_k \rVert^2)^{-1}} \hspace{1.5cm}(12)
\]

The goal of t-SNE optimization is to minimize the KL divergence between the distributions. $ P = \{ p_{ij} \} $ and $ Q = \{ q_{ij} \} $, which is given by:

\[
C = \text{KL}(P || Q) = \sum_i \sum_j p_{ij} \log \frac{p_{ij}}{q_{ij}} \hspace{1.5cm}(13)
\]

The algorithm iteratively adjusts the low-dimensional representations $ Y = \{ y_i \} $ to minimize this cost function by employing gradient descent. By exposing underlying patterns or clusters, this procedure aids in maintaining the local structure of the data points in the low-dimensional space.

\begin{flushleft}
    \textbf{3.3.4. Uniform Manifold Approximation and Projection}
\end{flushleft}

UMAP effectively extracts local and global structures from high-dimensional data while providing better data relationship preservation in lower-dimensional spaces (McInnes et al., 2018). After LDA, FA, and t-SNE, we presented a unique computational step wherein we configured a three-component UMAP model, which differs from the parameterizations used in the previous methods. UMAP sought to maximize the preservation based on the ideas of manifold learning. In contrast to the linear separability emphasis of LDA, the latent variable extraction of FA, or the probabilistic modeling of t-SNE, which focuses on local structures, UMAP used a topological approach, giving complex manifold relationships within the data priority. This customized method produced an improved three-dimensional model that provided a fresh viewpoint different from previous methods.
The objective function minimized by UMAP is given by:

{\tiny
\[
\underset{Y}{\text{arg min}}\left( \sum_{i=1}^{N} \sum_{j=1}^{N} f(d_{ij})\log \left(\frac{f(d_{ij})}{f(\hat{d}_{ij})}\right) + \lambda \sum_{i=1}^{N} \sum_{j=1}^{N} (1-w_{ij})(\hat{d}_{ij}-d_{ij})^{2} \right) \hspace{0.5cm}(14)
\]
}

Where:

- $Y$ represents the low-dimensional embedding that UMAP tries to optimize.

- $N$ denotes the number of data points or instances in the dataset.

- $f()$ is a function that operates on pairwise distances $d_{ij}$ between data points $i$ and $j$ in the original high-dimensional space.

- $d_{ij}$ and $\hat{d}_{ij}$ are the actual and approximated distances, respectively, between data points $i$ and $j$ in the high-dimensional space.

- $\lambda$ is a parameter that controls the trade-off between preserving local and global structure in the lower-dimensional embedding.

-$w_{ij}$ represents weights associated with the pairwise distances between data points
\subsection{3.4. Feature Integration}
\begin{flushleft}
    \textbf{3.4.1. Integration}  
\end{flushleft}
\hspace{0.5cm} We intentionally diverged from the proposal put forth by Vaidya and Vaidya (2022), which recommended replacing outliers following dimensionality reduction. This choice was carefully considered in light of the need to preserve data integrity and lessen the effect of outliers, particularly when employing t-SNE. t-SNE is intrinsically sensitive to outliers because it seeks to preserve local structures in the data. By giving outlier replacement precedence over dimensionality reduction, the intrinsic structure of the data is faithfully captured in the lower-dimensional space by taking care to shield data points from the perturbing effects of extreme values.

By reducing the divergence between probability distributions, the t-SNE algorithm highlights the similarities between nearby points and largely ignores the global structure. According to Halladin-Dabrowska et al. (2019), outliers can significantly impact the local structure and result in skewed representations in the lower-dimensional space by drawing or repelling neighboring points. The t-SNE cost function uses pairwise similarities, and the extreme values of outliers can disproportionately impact the optimization process and have a discernible impact on the visualization that is generated. The t-SNE cost function is defined by the Kullback-Leibler divergence between the conditional probabilities of pairs of data points in the high-dimensional space and their equal pairs in the low-dimensional space (van der Maaten et al., 2008). This can be expressed mathematically as equation (13). In contrast, LDA searches for a lower-dimensional space where the maximum separation between different classes exists. LDA is less prone to outliers since it is a linear method that focuses on the global structure and class separability. Outliers have less of an impact on the determination of the discriminant axes because LDA is based on the overall distribution of data points and their class labels. Class separability in the feature space is emphasized by the LDA's objective.

\begin{flushleft}
    \textbf{3.4.2. Scaling}
\end{flushleft}

The \texttt{StandardScaler} was used to normalize the dataset's features. A common technique used to make sure features are on a similar scale is standard scaling. In the underlying mathematics, each feature is transformed independently by dividing by its standard deviation and subtracting its mean. In mathematical terms, the transformation for a feature $X_i$ in the dataset is $X_i' = \frac{X_i - \text{mean}(X_i)}{\text{std}(X_i)} \hspace{1.5cm}(15)$.

The scaling guarantees that all features are on an equalized scale, which is important for models that depend on distance metrics (e.g., support vector machines or k-nearest neighbors) or other features that are sensitive to feature scales. By taking the mean out of every feature, it accomplishes mean centering, which makes features easier to compare and understand. Additionally, \texttt{StandardScaler} maintains the data's distributional shape, which makes it a good option when the analysis depends on preserving the original distribution. Although there are other scaling techniques such as Robust and Min-Max scaling, \texttt{StandardScaler} is especially suitable in cases where the features have a distribution that is roughly normal.

\subsection{3.5. Model Assessment}
\begin{flushleft}
    \textbf{3.5.1. Train-Test Split}
\end{flushleft}

We used stratified splitting to maintain class proportions when separating the dataset into training (75\%) and testing (25\%) subsets. Reproducibility was guaranteed by the random state of 42. This division made it possible to train and assess models independently for an objective appraisal of performance.

\begin{flushleft}
    \textbf{3.5.2. Cross Validation}
\end{flushleft}

10-fold cross-validation was also used to assess the of the model. Using this method, the dataset was divided into 10 subsets, 9 of which were used for training and one for validation, alternately.

\section{4. Results and Discussion}
\hspace{0.5cm}We used multiple dimensionality reduction techniques on our dataset and sought to compress the feature space: LDA concentrated on class separability, FA captured latent variables, t-SNE prioritized local structures,  and UMAP preserved both local and global structures in lower dimensions. The ensuing discourse contrasts and evaluates these methodologies, accentuating their individual merits and demerits in proficiently portraying intricate data structures. The algorithms used were LR, MLP, KNN and RF.

\subsection{4.1. Visualization and Evaluation}
\begin{flushleft}
    \textbf{4.1.1. ROC Curves}
\end{flushleft}

\begin{figure}[h] 
    \centering
    \includegraphics[width=1\linewidth]{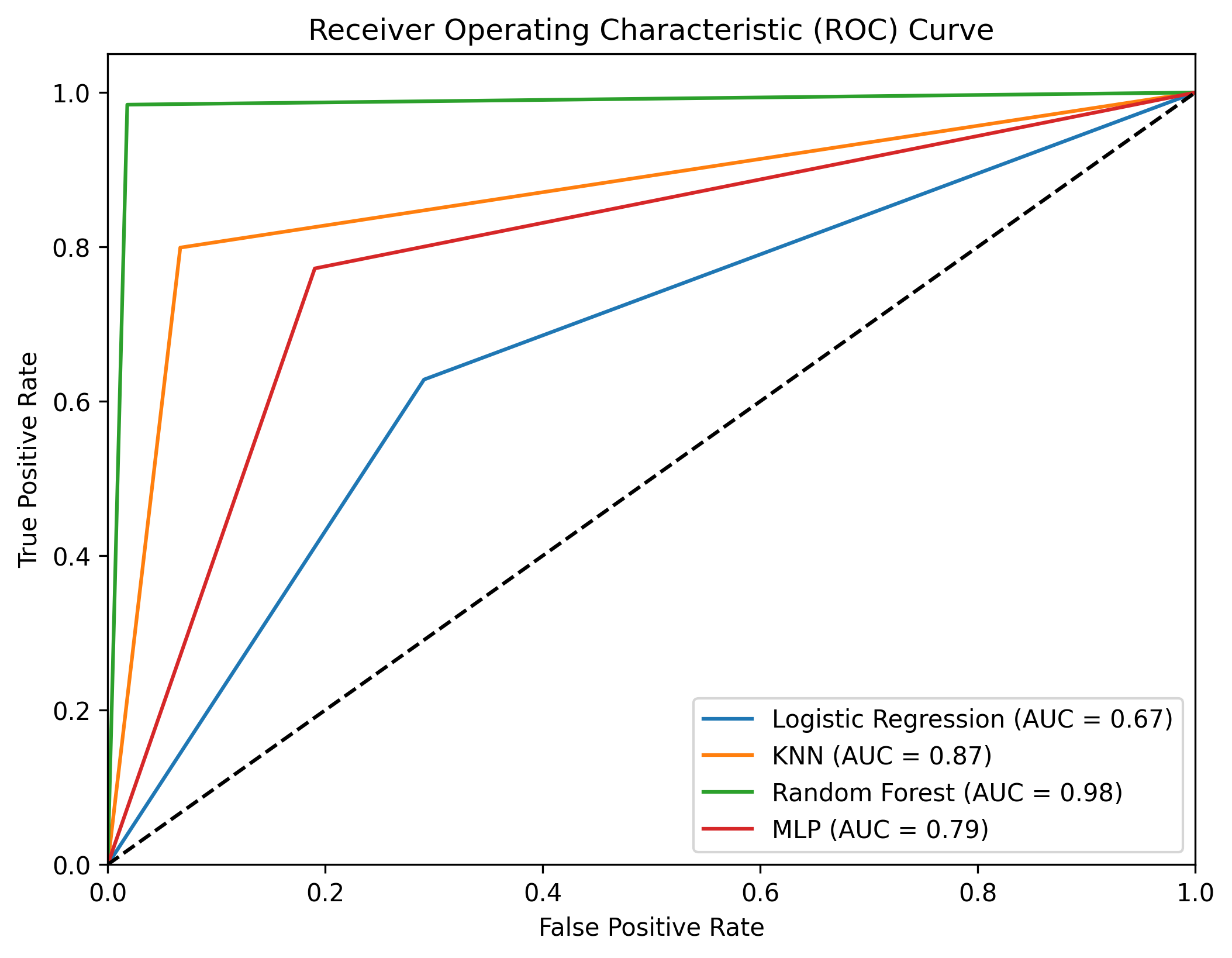}
    \caption{ROC curves in 10-fold CV}
    \label{fig:ROC Curve}
\end{figure}
\begin{figure}[h]
    \centering
    \includegraphics[width=1\linewidth]{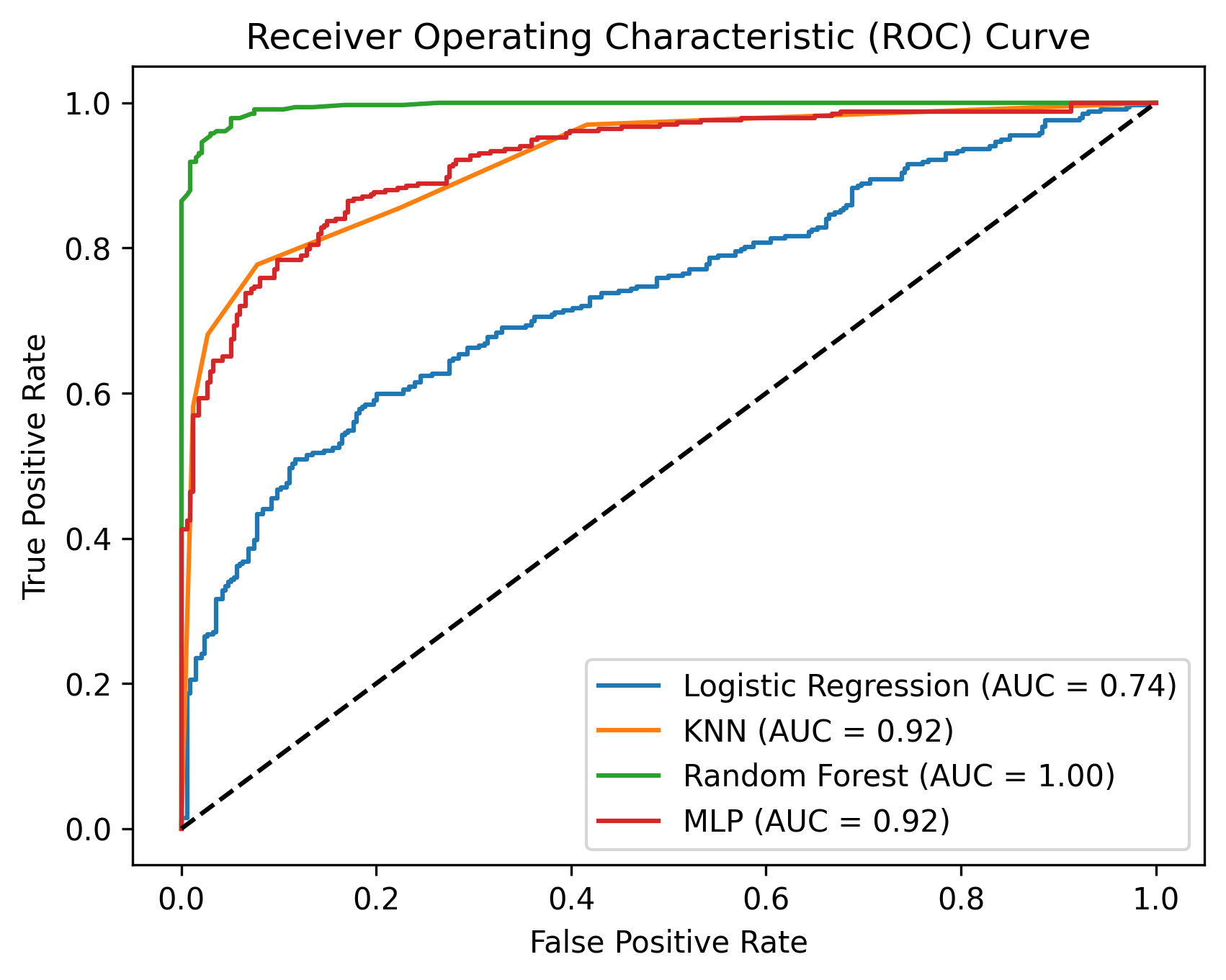}
    \caption{ROC curves in Train-Test Split}
    \label{fig:ROC Curve}
\end{figure}
\hspace{0.5cm}The ability of each model to differentiate between classes was clearly shown by figures 2 and 3. Above all False Positive Rate thresholds, Random Forest maintained higher True Positive Rates than the others. KNN and MLP showed moderate discriminatory abilities, while Logistic Regression exhibited relatively lower discriminatory power. 

\begin{flushleft}
    \textbf{4.1.2. Precision-Recall Curves}
\end{flushleft}

The curve for precision and recall for each algorithm was figures 4 and 5. Across a range of recall thresholds, Random Forest and Multi-Layer Perceptron demonstrated superior precision levels and outperformed Logistic Regression and K-Nearest Neighbors.
\begin{figure}[h] 
    \centering
    \includegraphics[width=1\linewidth]{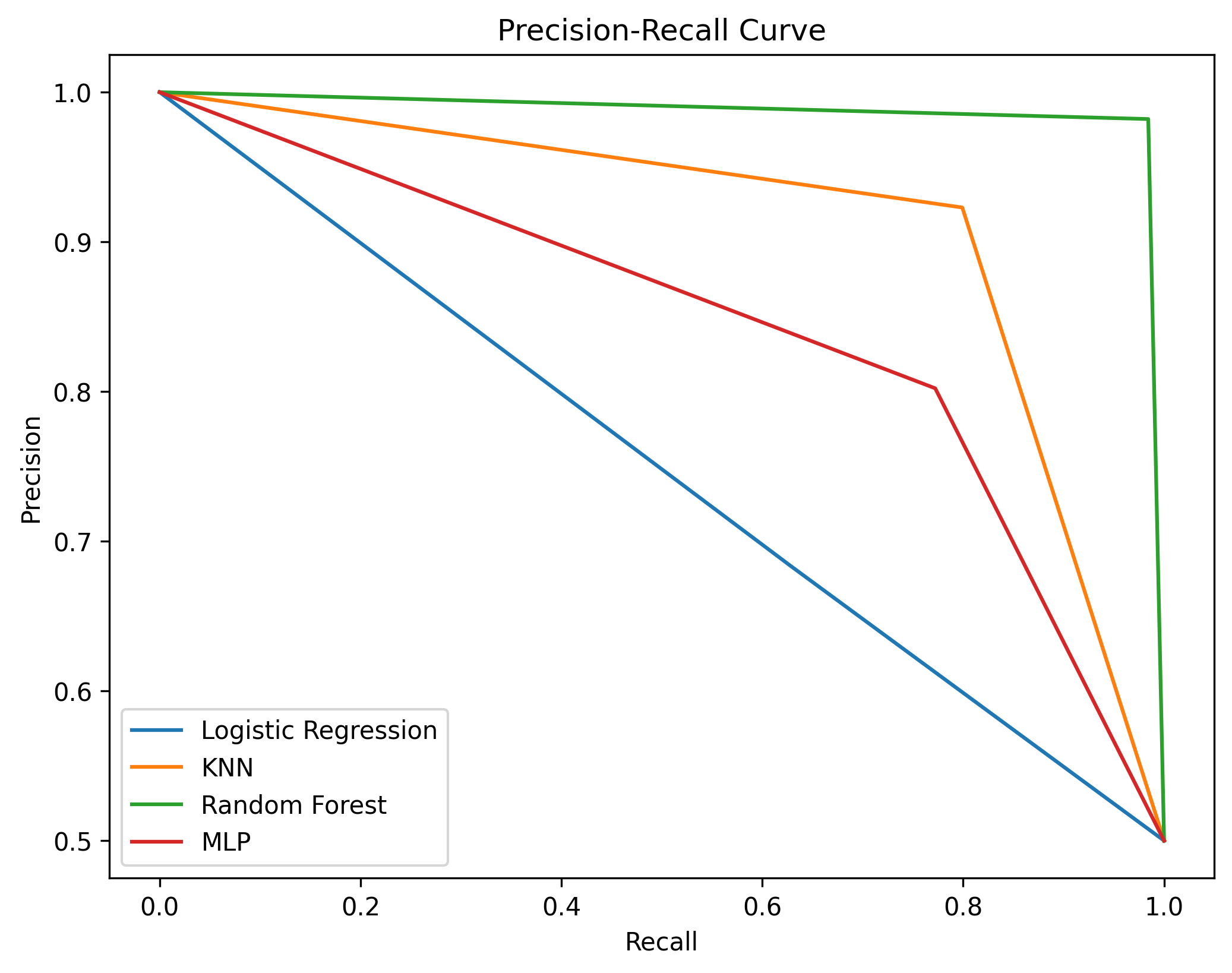}
    \caption{Precision-Recall curves in 10-fold cross validation}
    \label{fig:Precision Recall Curve}
\end{figure}
\begin{figure}[h] 
    \centering
    \includegraphics[width=1\linewidth]{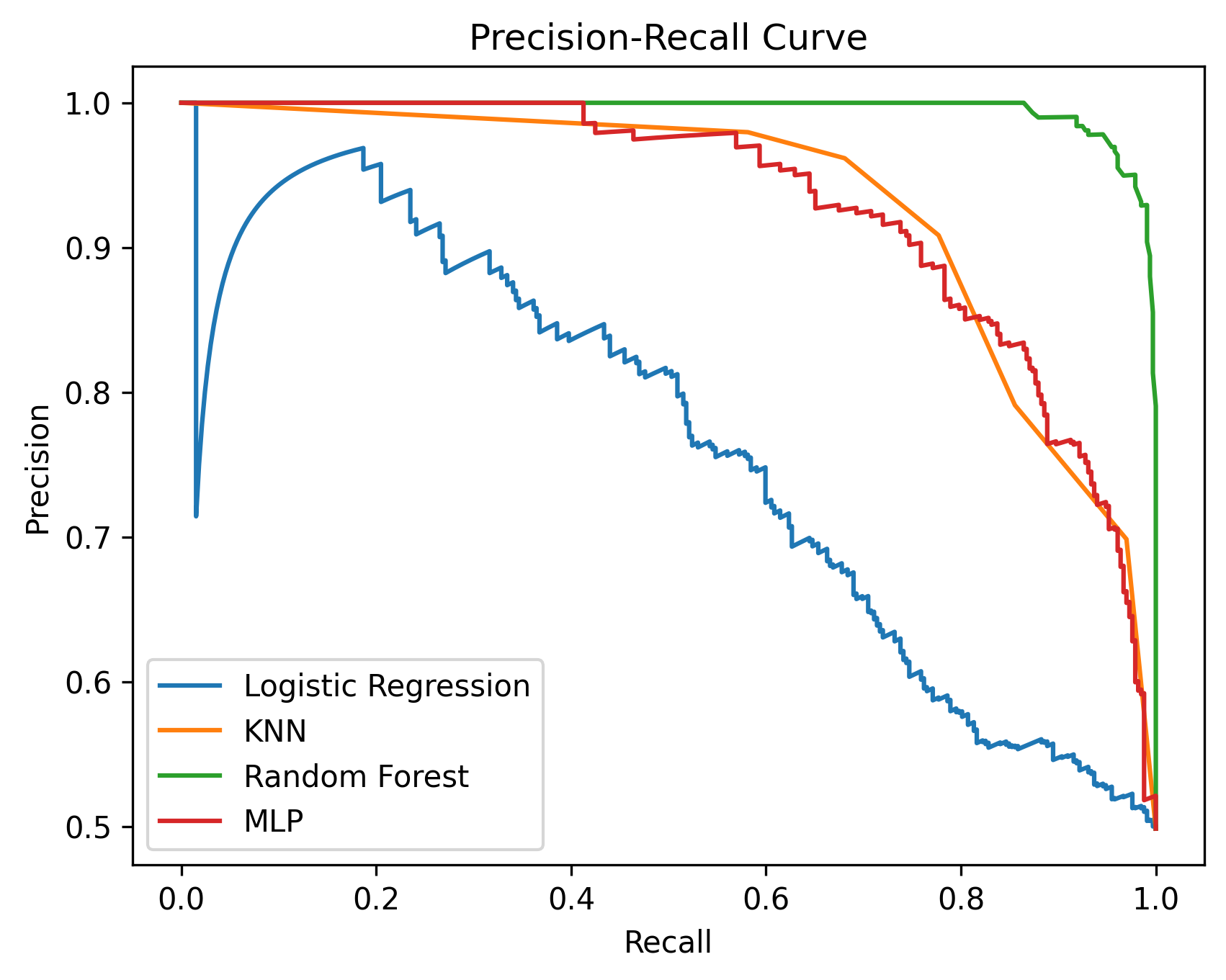}
    \caption{Precision-Recall curves in Train-Test Split}
    \label{fig:PrecisionRecall Curve}
\end{figure}

Random Forest demonstrated its superior ability to retrieve relevant instances without sacrificing precision in the train-test split analysis, as evidenced by its ability to maintain higher precision levels across a range of recall thresholds. While KNN demonstrated remarkable precision but marginally worse recall performance, MLP demonstrated balanced but moderate performance whereas Logistic Regression fell out.

\begin{flushleft}
    \textbf{4.1.3. Evaluation Metrics Bar chart}
\end{flushleft}
\begin{figure*}[h]
  \centering
  \includegraphics[width=0.8\textwidth]{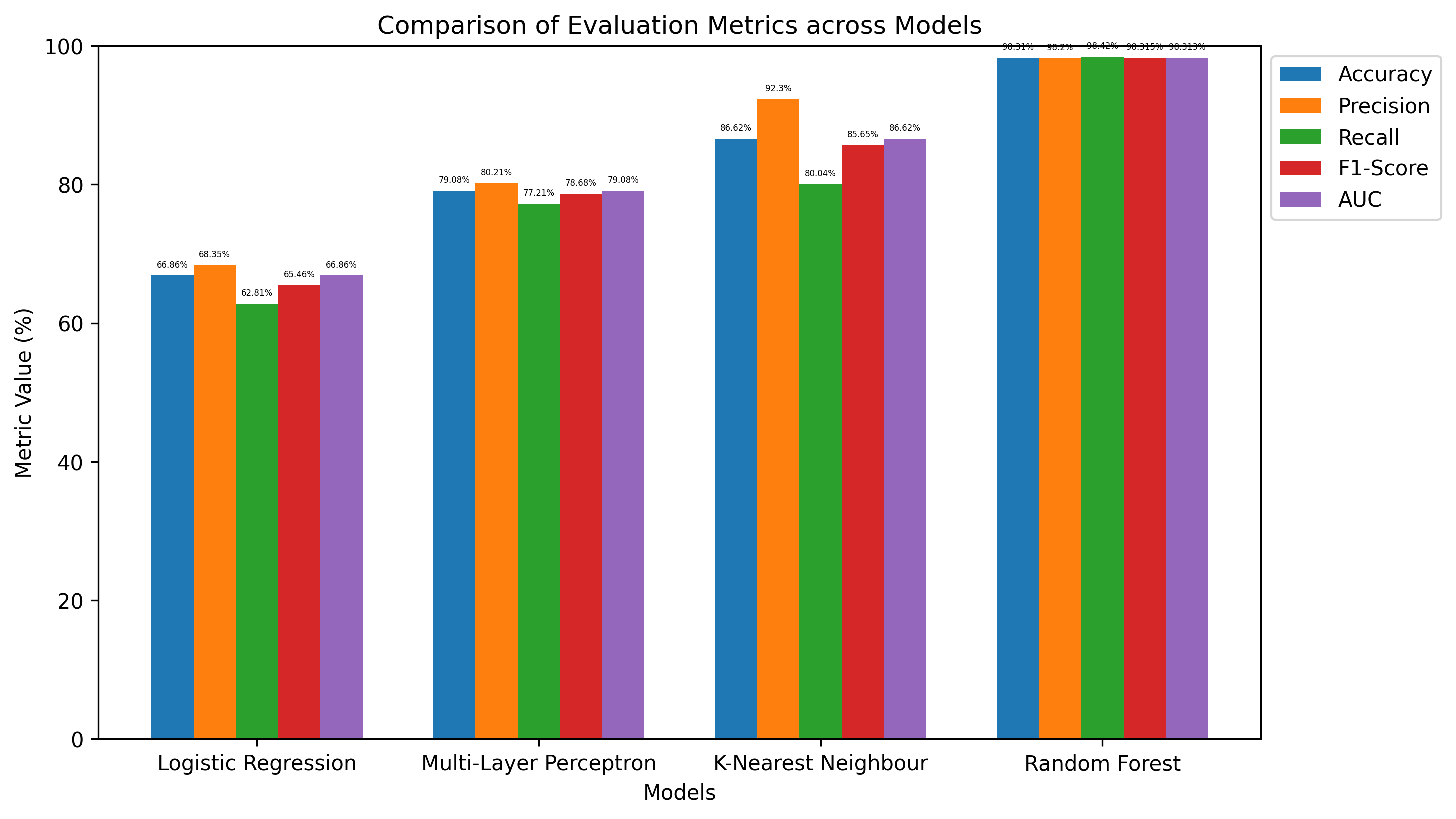} 
  \caption{Bar chart representing evaluation metrics for each algorithm in 10-fold cross validation}
  \label{fig:full-width}
\end{figure*}

Evaluation metrics were used to gauge model performance, depicted in a bar chart represented in figures 6 and 7. Random Forest led with 98.31\% accuracy, while KNN followed at 86.62\%. Recall ranged from 62.81\% (Logistic Regression) to 98.42\% (Random Forest), with precision at 98.20\% (Random Forest) and 92.30\% (KNN). In cross-validation, Random Forest peaked at 98.315\% F1-Score, matching its AUC and accuracy.
\begin{figure*}
  \centering
  \includegraphics[width=0.8\textwidth]{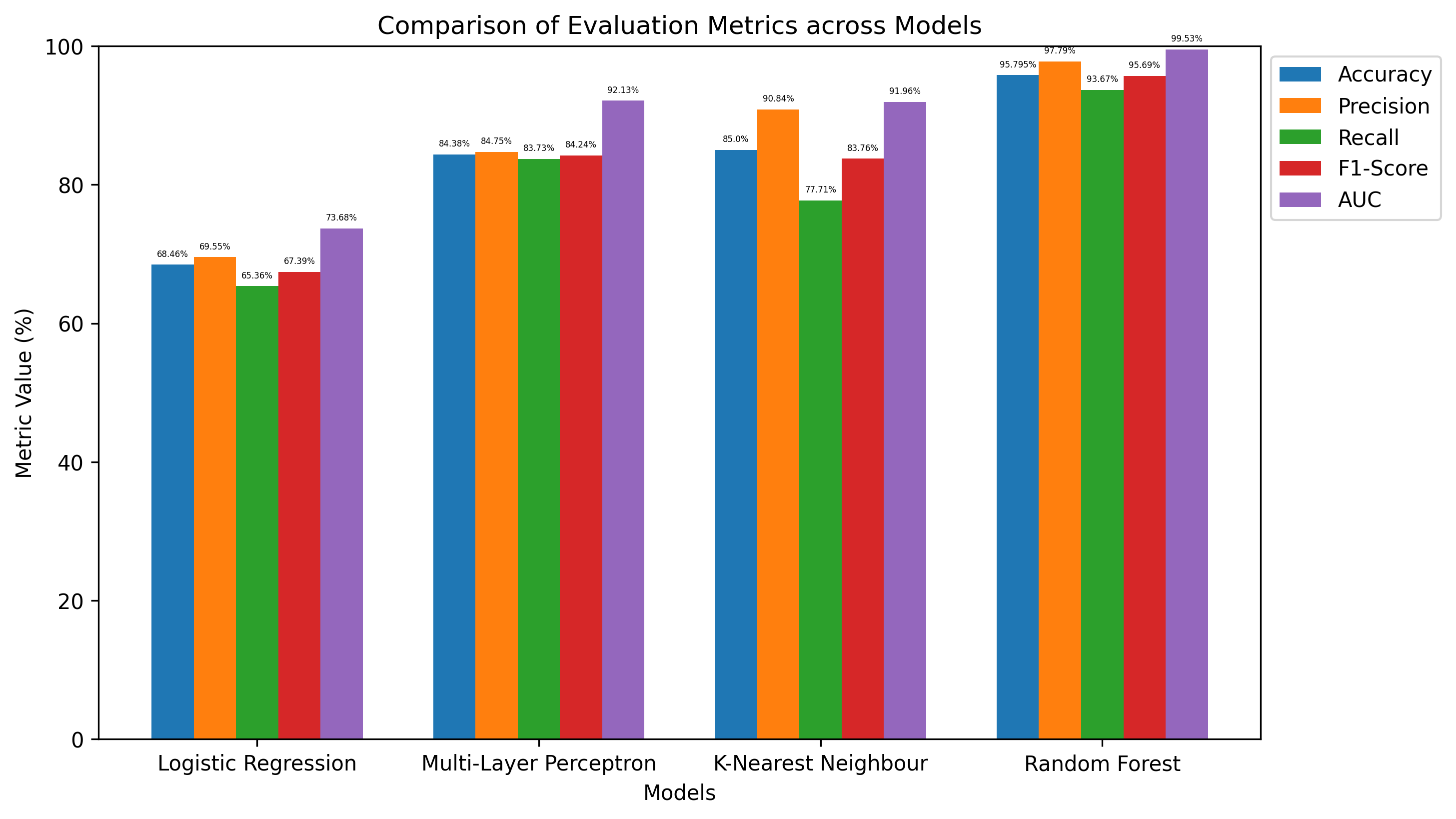} 
  \caption{Bar chart representing evaluation metrics for each algorithm in Train-Test Split}
  \label{fig:full-width}
\end{figure*}
In the train-test split analysis, Random Forest excelled across all metrics: 95.79\% accuracy, 97.79\% precision, and 99.53\% AUC. KNN followed closely with strong precision (90.84\%) and accuracy (85\%). MLP showed moderate, balanced predictive capabilities whereas Logistic Regression was lagging behind.

\begin{flushleft}
    \textbf{4.1.4. Runtime Bar Chart}
\end{flushleft}

\begin{figure}
    \centering
    \includegraphics[width=\linewidth]{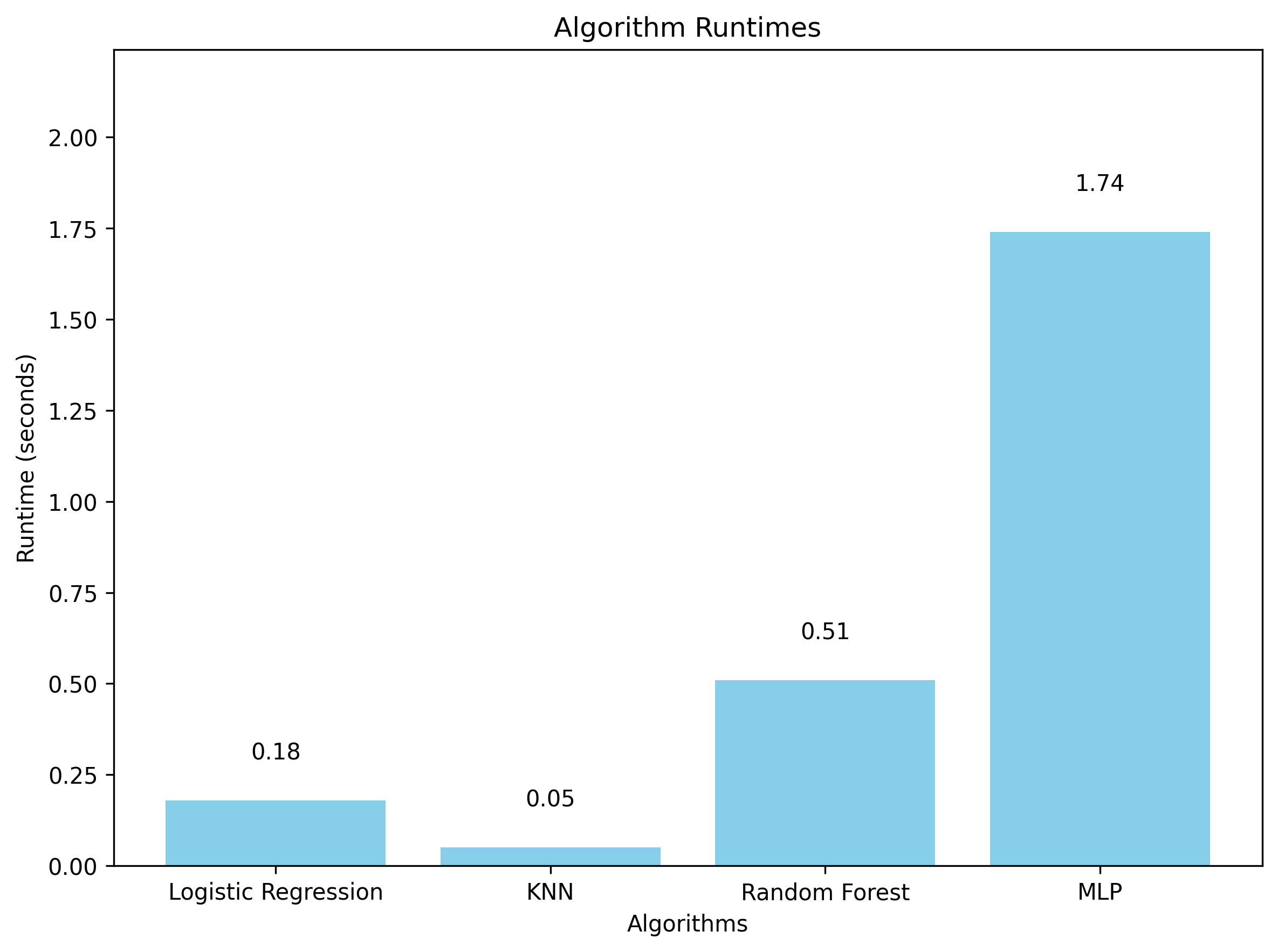}
    \caption{Runtime of different algorithms}
    \label{fig:Runtime Curve}
\end{figure}

According to the runtime analysis, the models' computational requirements varied, with KNN requiring the least amount of time (0.05 seconds) and MLP requiring the longest (1.74 seconds) as shown in the figure 8. This difference emphasizes how crucial it is to take computational efficiency into account when choosing a model. Our study was carried out on a system that had an Intel Core i-7 with 16 GB RAM and a 10th generation CPU, and it provided insight into how well the model performed in different computing environments. Making informed decisions about which models to use for specific computational configurations is made easier by this short runtime curve.

\begin{flushleft}
    \textbf{4.1.5. Calibration Curve}
\end{flushleft}

Brier Scores, which measure the precision of predicted probabilities, were carefully used to evaluate the calibration efficiency of the models. The calibration of the logistic regression was deemed acceptable, as evidenced by its moderate Brier Score of 0.2028. In contrast, with a score of 0.1113, KNN demonstrated better calibration. Among them, Random Forest performed the best out of all of them as shown in figure 9, with a considerably lower Brier Score of 0.0351, indicating higher probability prediction accuracy. Comparable to KNN in terms of performance was the MLP, with a Brier Score of 0.1157.

\begin{figure}[h]
    \centering
    \includegraphics[width=\linewidth]{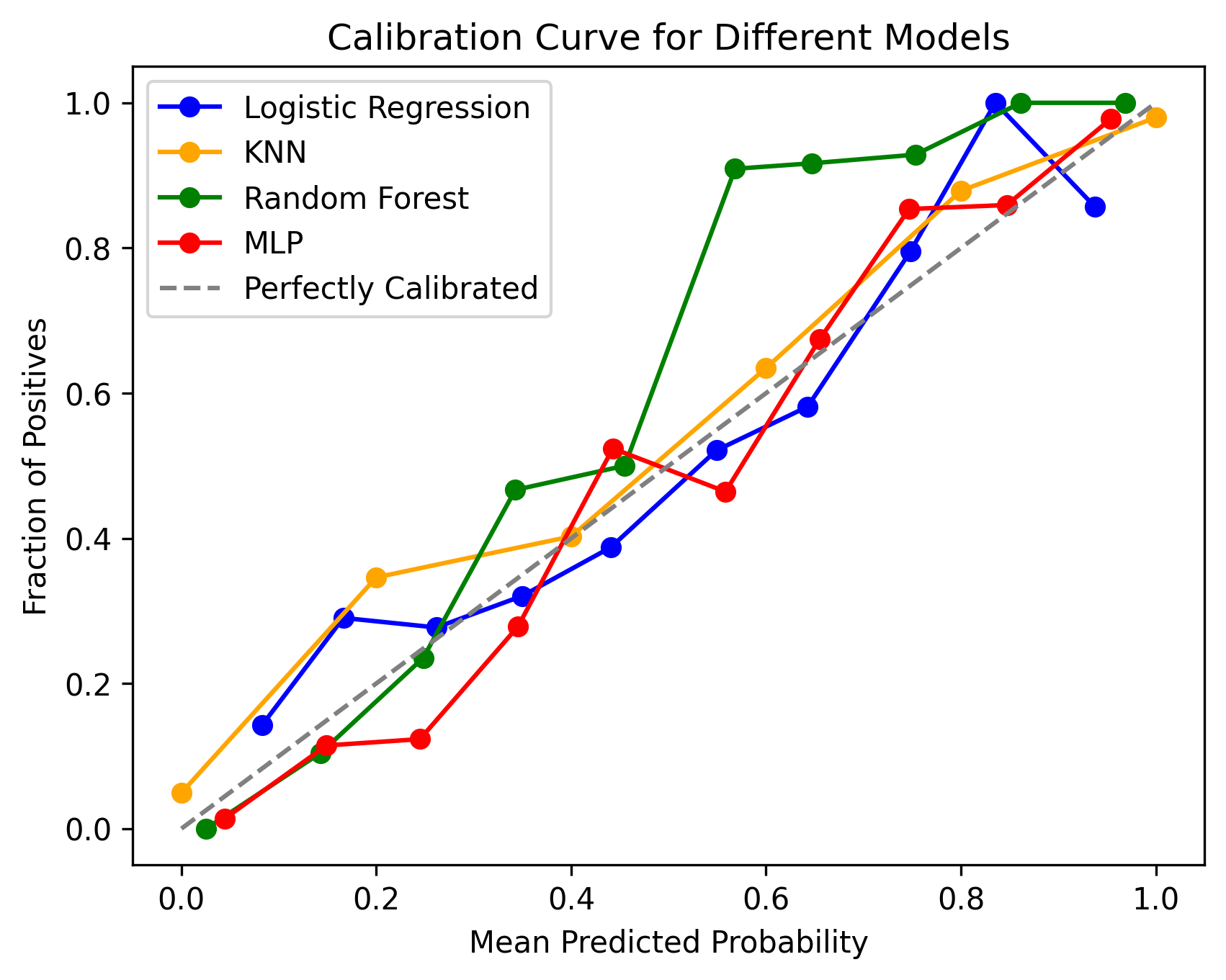}
    \caption{Learning curve for different algorithms in train-test split}
    \label{fig:CC}
\end{figure}

The use of a calibration curve is justified by its capacity to provide information about model reliability by illuminating the relationship between expected probabilities and actual results. Isotonic Regression refines the accuracy of probability estimates by transforming predicted probabilities to closely match observed frequencies, thus improving calibration even further. Interestingly, Random Forest turns out to be the best calibration technique, outperforming MLP, KNN, and Logistic Regression. This assessment provides important information about the models' ability to produce accurate probability estimates, which is crucial for applications that depend on well-calibrated forecasts. The results of the study highlight the need of using reliable calibration methods, with Random Forest demonstrating particularly good results.

\begin{flushleft}
    \textbf{4.1.6. Learning Curve for Random Forest}
\end{flushleft}

Analyzing learning curves to investigate the Random Forest classifier's performance revealed fascinating dynamics as per figures 10 and 11. The training score, represented by the blue line, demonstrated impressive stability, indicating a constant high accuracy over the course of exposure to the training dataset. By comparison, the cross-validation (CV) score (orange line) started at around 76\% and rises steadily as the training dataset grows to 95.79\%, suggesting improved generalization. The CV score's upward trajectory indicated that the model skillfully integrated more training data to improve its predictive abilities on cases that had not yet been observed.

\begin{figure}[h]
    \centering
    \includegraphics[width=\linewidth]{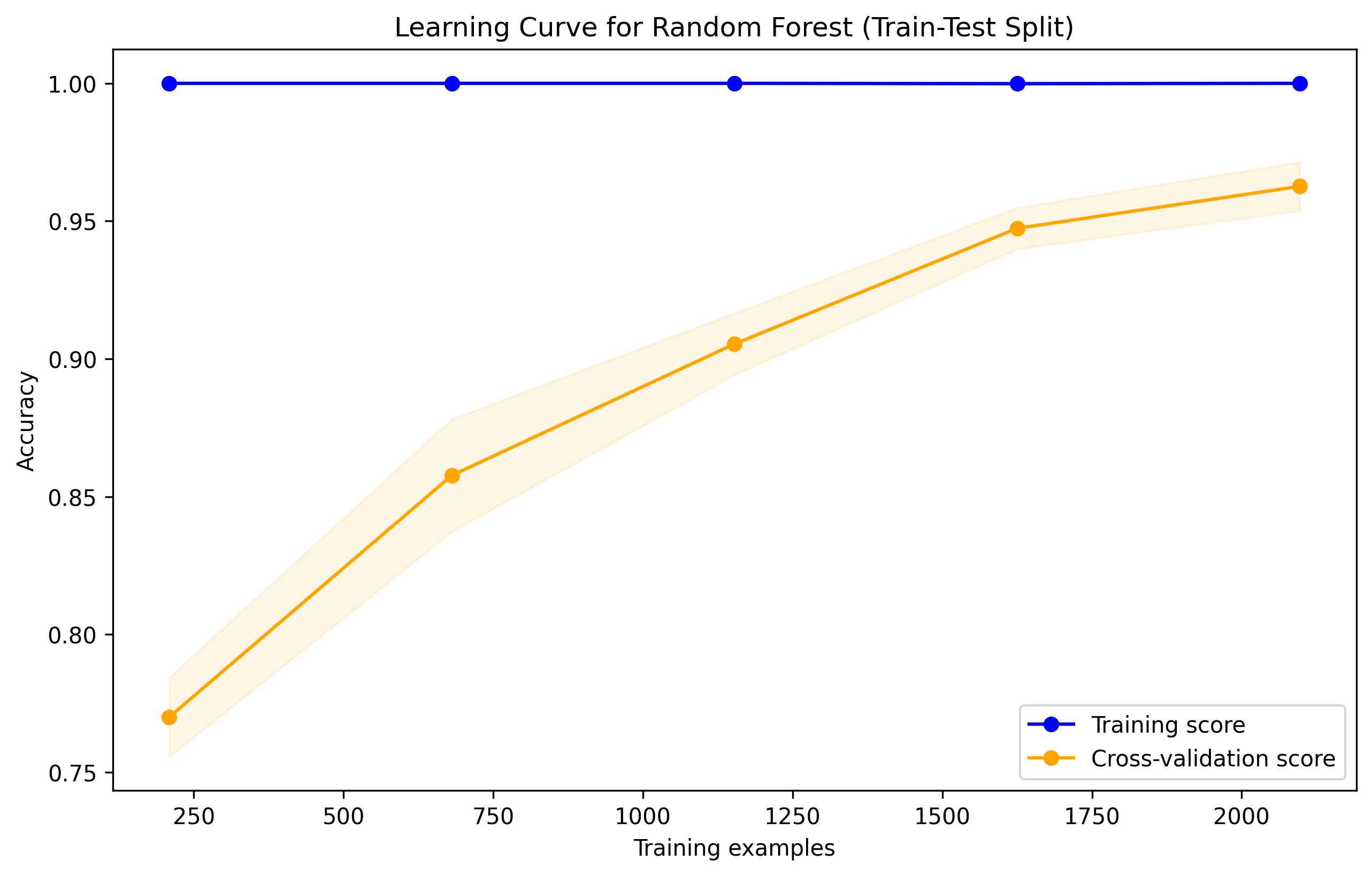}
    \caption{Learning curve for random forest in train-test split}
    \label{fig:LCRFT}
\end{figure}
\begin{figure}[h]
    \centering
    \includegraphics[width=\linewidth]{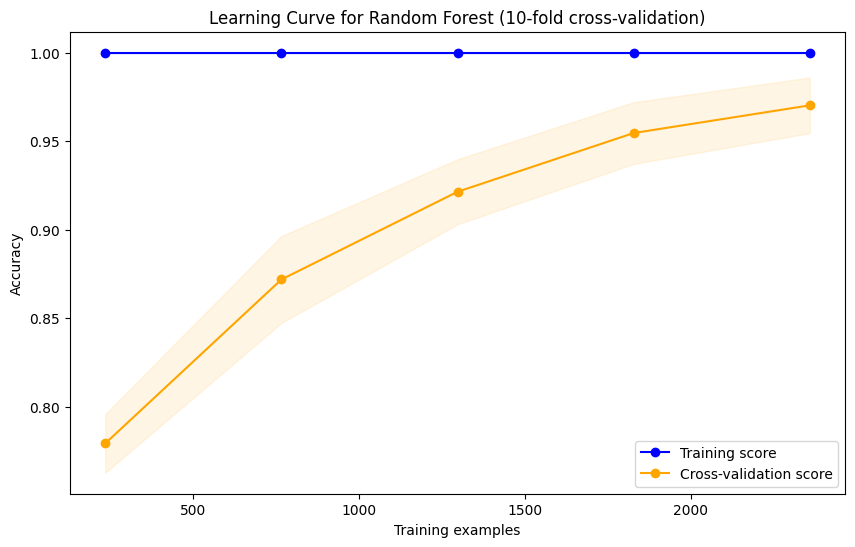}
    \caption{Learning curve for random forest in 10-fold cross validation}
    \label{fig:LCRFT}
\end{figure}

The CV score increased dramatically to 98.31\% during the learning curve plotting with a 10-fold cross-validation, starting at about 77\%. This additional knowledge strengthens the model's robustness by highlighting its ability to generalize across various folds. A well-generalized model is implied by the training score's consistent stability and the CV score's convergence towards it. The decreasing difference between the two scores is noteworthy because it highlights the model's resistance to overfitting. The model is especially well-suited for the task at hand because of its balanced learning process, which places it in an adaptive position to handle different training dataset sizes. In conclusion, the learning curve story illustrated a Random Forest model that balanced stability and flexibility, demonstrating complex learning dynamics for successful generalization.

\begin{flushleft}
    \textbf{4.1.7. Confusion Matrix for Random Forest}
\end{flushleft}

When assessing the RF model for disease prediction, the use of a confusion matrix produced informative results. The matrix, in figure 12, demonstrated the model's ability to correctly identify 313 cases of liver disease (LD) and 329 cases of non-liver disease (NLD). In addition to exposing five instances of misclassifying NLD as LD and 19 cases of misclassifying LD as NLD, this impressive performance was tainted by errors. A confusion matrix, which provides a thorough breakdown of true positives (TP), true negatives (TN), false positives (FP), and false negatives (FN), was crucial in conducting a thorough evaluation of the model's predictive abilities. 

\begin{figure}[h]
    \centering
    \includegraphics[width=\linewidth]{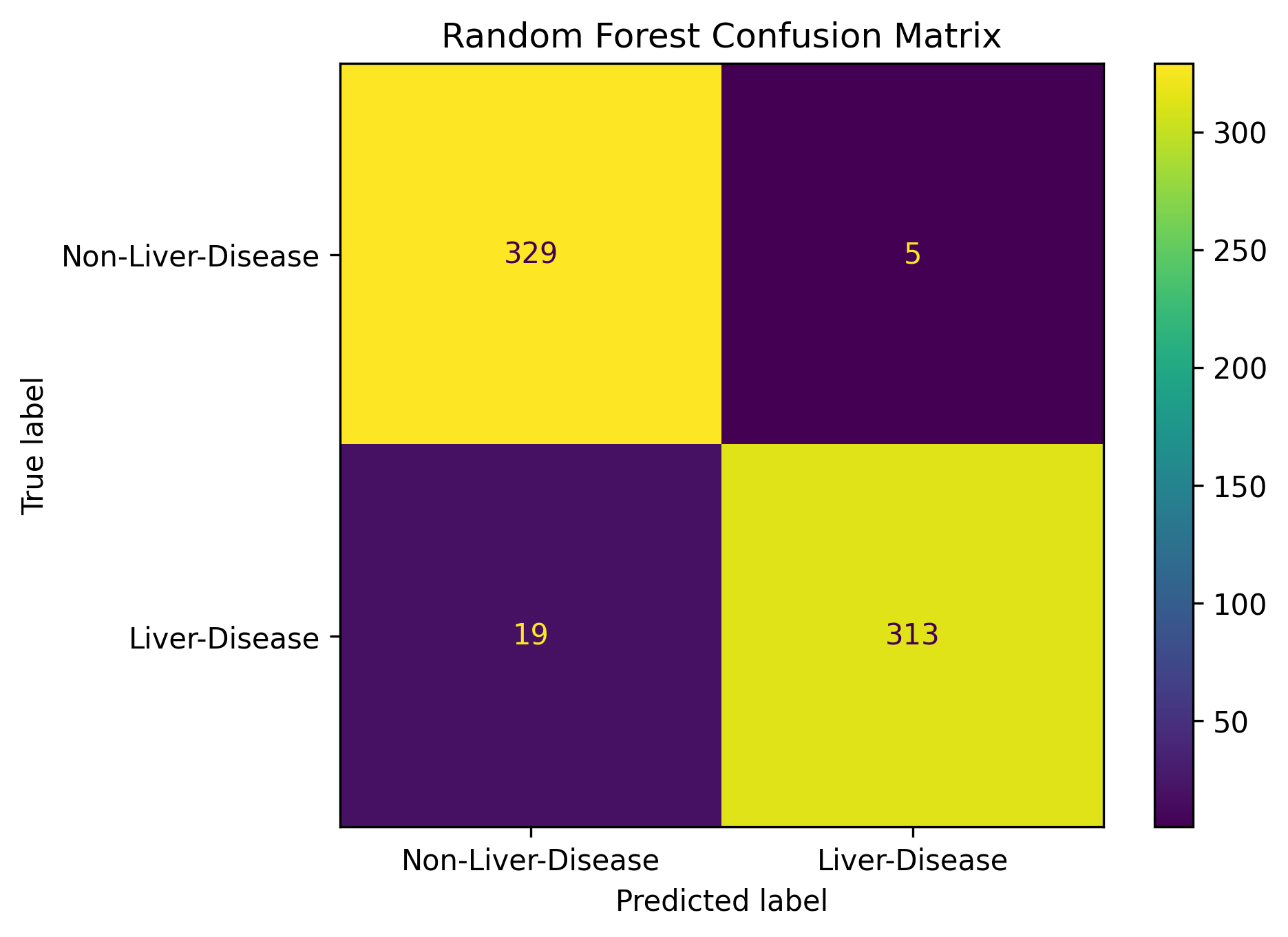}
    \caption{Confusion matrix for random forest in train-test split}
    \label{fig:cm}
\end{figure}

These results not only showed that the model could produce accurate predictions, but they also highlighted some areas that still required work. The confusion matrix was a useful tool for a more accurate and nuanced assessment of the RF model's performance in liver disease prediction because it presented these nuanced metrics. 

\subsection{4.2. Performance of different algorithms}
\begin{table}[h]
\centering
\caption{Performance of different algorithms in 10-fold cross validation}
\label{table:eval_metrics}
\resizebox{\columnwidth}{!}{%
\begin{tabular}{|p{1.5cm}|p{1.5cm}|p{1.5cm}|p{1.5cm}|p{1.5cm}|p{1.5cm}|}
\hline
\textbf{Model} & \textbf{Accuracy (\%)} & \textbf{Precision (\%)} & \textbf{Recall (\%)} & \textbf{F1-Score (\%)} & \textbf{AUC (\%)} \\
\hline
Logistic Regression & 66.86 & 68.35 & 62.81 & 65.46 & 66.86 \\
MLP & 79.08 & 80.21 & 77.21 & 78.68 & 79.08 \\
KNN & 86.62 & 92.30 & 80.04 & 85.65 & 86.62 \\
\textbf{Random Forest} & \textbf{98.31} & \textbf{98.20} & \textbf{98.42} & \textbf{98.315} & \textbf{98.313} \\
\hline
\end{tabular}%
}
\end{table}

\begin{table}[h]
\centering
\caption{Performance of different algorithms in train-test split}
\label{table:eval_metrics}
\resizebox{\columnwidth}{!}{%
\begin{tabular}{|p{1.5cm}|p{1.5cm}|p{1.5cm}|p{1.5cm}|p{1.5cm}|p{1.5cm}|}
\hline
\textbf{Model} & \textbf{Accuracy (\%)} & \textbf{Precision (\%)} & \textbf{Recall (\%)} & \textbf{F1-Score (\%)} & \textbf{AUC (\%)} \\
\hline
Logistic Regression & 68.46 & 69.55 & 65.36 & 67.39 & 73.68 \\
MLP & 84.38 & 84.75 & 83.73 & 84.24 & 92.13 \\
KNN & 85.00 & 90.84 & 77.71 & 83.76 & 91.96 \\
\textbf{Random Forest} & \textbf{95.79} & \textbf{97.79} & \textbf{93.67} & \textbf{95.69} & \textbf{99.53} \\
\hline
\end{tabular}%
}
\end{table}

\hspace{0.5cm}When we evaluated the models' performance using cross-validation and train-test splits, the Random Forest model consistently performed better in both cases as shown in tables 2 and 3. With a remarkable accuracy of 95.79\% in the train-test split, the Random Forest model outperformed Logistic Regression (68.46\%), KNN (85\%), and MLP (84.38\%). In the same way, the Random Forest continued to outperform the other models in cross-validation, with a mean accuracy of 98.31\%. Because of its consistent high performance, the Random Forest model is positioned as dependable and strong for the particular dataset, indicating its potential for precise predictions in real-world applications.

\begin{figure}[h]
    \centering
    \includegraphics[width=\linewidth]{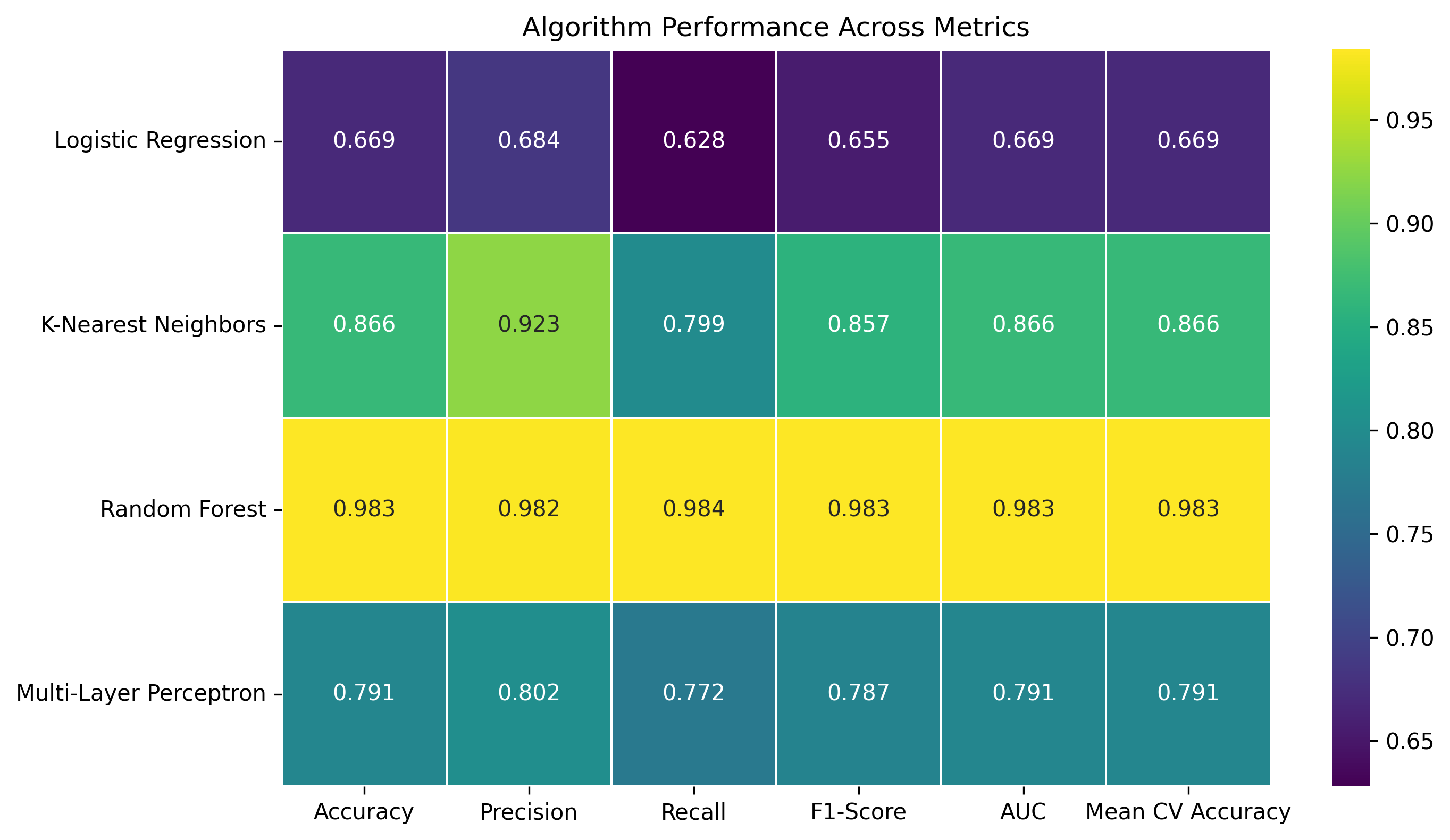}
    \caption{Heatmap for different algorithms in 10-fold cross validation}
    \label{fig:HM}
\end{figure}

The Random Forest's consistent performance in a variety of assessment methods highlights both its adaptability to generalize well on unknown data and its resistance to overfitting. Because of its resilience, it is a sensible option for predictive modeling in situations similar to the dataset under study.Our analysis's results highlight the Random Forest's potential for precise forecasts, promoting its use in real-world applications that demand strong predictive abilities.

\subsection{4.3. Comparative Analysis}
\hspace{0.5cm}In comparison to recent state-of-the-art studies, this comparative evaluation in table 4 highlights the superior performance of our proposed method on the ILPD dataset and demonstrates its efficacy in achieving high predictive accuracy. The table shows that our strategy—which used Random Forest (RF) in particular—achieved the highest accuracy of 98.31\%, outperforming all other approaches.
\begin{table}[h]
\centering
\caption{Performance comparison of recent state-of-the-art studies with proposed method on the ILPD dataset}
\label{table:eval_metrics}
\resizebox{\columnwidth}{!}{%
\begin{tabular}{|p{3 cm}|p{1cm}|p{1.8cm}|p{1.5cm}|p{1.3cm}|}
\hline
\textbf{Author(s)} &\textbf{Year}  & \textbf{Protocol} & \textbf{Classifiers} & \textbf{Accuracy (\%)} \\
\hline
Amin et al. & 2023 & Cross & RF & 88.10 \\
& & Validation & MLP & 83.53 \\
& & & Ensemble & 82.09 \\
& & & KNN & 67.90 \\
& & & SVM & 67.90 \\
& & & LR & 55.40 \\
\hline
Niha et al. & 2023 & Train-test & LR & 76.575 \\
&& Split & RF & 74.28 \\
&&& KNN & 70.20 \\
\hline
Anthonysamy & 2023 & Train-test & HVC & 78.62 \\
and Babu & & Split & MLP & 77.24 \\
& & & SVM & 76.55 \\
& & & KNN & 73.10 \\
\hline
Bhupathi et al. & 2022 & CRISP-DM & Auto encoders & 92.1 \\
& &  & KNN & 91.7 \\
& & & CART & 83.6 \\
& & & SVM & 78.1 \\
& & & LDA & 70.9 \\
& & & Naive Bayes & 65.1 \\
\hline
Sivasangari et al. & 2022 & Train-test & SVM & 95.18 \\
 & & Split, Cross & RF & 92.77 \\
& & Validation & DT & 87.95 \\
\hline
Ghosh et al. & 2021 & Train-test & RF & 83.76 \\
& & Split & XGBoost & 82.05 \\
& & & SVM & 81.20 \\
& & & KNN & 81.20 \\
& & & LR & 79.49 \\
& & & DT & 79.49 \\
& & & AdaBoost & 76.07 \\
\hline
Panwar et al. & 2021 & Train-test & SVM & 74.09 \\
& & Split & DT & 72.54 \\
& & & RF & 72.02 \\
& & & LR & 71.50 \\
& & & Naive Bayes & 57.51 \\
\hline
Kuzhippallil et al. & 2020 & Genetic & RF & 88 \\
& & Algorithm & XGBoost & 86 \\
& & & Light GBM & 86 \\
& & & SE & 85 \\
& & & DT & 84 \\
& & & GB & 84 \\
& & & AdaBoost & 83 \\
& & & MLP & 82 \\
& & & KNN & 79 \\
& & & LR & 76 \\
\hline
\textbf{Proposed }& & \textbf{Cross} & \textbf{RF} & \textbf{98.31} \\
\textbf{Method} & & \textbf{Validation} & KNN & 86.62 \\
& & & MLP & 79.08 \\
& & & LR & 66.86 \\
\hline
\end{tabular}%
}
\end{table}
Interestingly, KNN and MLP also showed significant accuracy at 86.62\% and 79.08\%, respectively, proving the effectiveness of our approach. 

RF demonstrated the highest accuracy at 88.10\% when used in a cross-validation protocol by Amin et al. (2023), followed by MLP at 83.53\%. Niha et al. (2023) examined RF, KNN and LR in train-test split. LR obtained an 87.16\% F1 score, 76.57\% accuracy, 75.00\% precision, and 99.50\% recall. KNN showed 84.32\% F1 score, 84.90\% recall, 83.10\% precision, and 70.20\% accuracy. RF displayed an 80.32\% F1 score, 77.90\% recall, 84.40\% precision, and 74.28\% accuracy. The study evaluated TP, TN, FP, and FN rates with a focus on liver disease classification. These outcomes demonstrated the effectiveness of the algorithms and offered insightful information about how they might be used for machine learning tasks involving classification and predictive modeling. 

In order to predict liver disease, Anthonysamy and Babu (2023) used machine learning algorithms such as RF, KNN, and LR. Scaling, column elimination, and KNN-imputer for null values were all part of the data preprocessing. The Hard Voting Classifier (HVC) and MLP were introduced. HVC came out to be better than other classifiers, according to the results, with 78.62\% accuracy, 80\% specificity, 87\% F-score, 78\% precision, and 78\% recall. 77\% accuracy, 100\% specificity, 87\% F-Score, 77\% precision, and 100\% recall were attained by MLP. 96\% recall, 76\% accuracy, and 96\% specificity were attained by SVM using an RBF kernel. KNN showed 89\% recall, 73\% accuracy, and 89\% specificity. Bhupathi et al. (2022) carried out a study in 2022 to evaluate different classifiers in compliance with the CRISP-DM (Cross-Industry Standard Process for Data Mining) protocol. Notably, their results showed that autoencoders, with an astounding accuracy score of 92.1\%, emerged as the most accurate classifier. Using a train-test split and cross-validation, Sivasangari et al. (2022) found that SVM had the highest accuracy, at 95.18\%. Ghosh et al. (2021) investigated liver disease prediction using RF, KNN, SVM, Decision Tree, and other ensemble methods. The methodology included feature scaling as a necessary preprocessing step prior to training datasets with sizes ranging from 50\% to 90\%. RF was clearly the most successful algorithm, with accuracy of 83.76\%, precision of 87\%, recall of 93.5\%, and an F1 score of 90.1\%. RF's AUC was 81.3\%.

Panwar et al. (2021) used advanced pre-processing techniques, such as replacing missing values with null values and their corresponding instances, in order to detect chronic liver disease. An approach that combined filter and wrapper techniques was used to carefully select features. Through correlation analysis, attributes with a correlation of greater than 70\% were first disregarded. Training (70\%) and test (30\%) sets were created from the dataset through randomization. SVM achieved an impressive accuracy of 74.09\%. Various classification algorithms were implemented, including Naïve Bayes, LR, RF, and Support Vector Machine. Kuzhipallil et al. (2020) looked at preliminary data analysis, using data pre-processing methods like imputation, label encoding, duplicate value removal, resampling, and outlier detection, as well as data exploration to summarize and visualize patterns. Feature selection was made easier by genetic algorithms, which optimized input features for predictive models. Several machine learning algorithms, including shortened versions like RF and XGBoost, were used for classification. LightGBM and Stacking Estimator showed the highest accuracy (88\%) in the study, highlighting enhanced performance following feature selection and outlier removal. 

Using a cross-validation protocol, proposed method outperformed all other studies with RF obtaining an astounding accuracy of 98.31\%. Significantly, KNN and MLP also showed good accuracy at 86.62\% and 79.08\%, respectively, proving the effectiveness of the suggested methodology.
\section{5. Conclusion}

\hspace{0.5cm}Our thorough investigation of ML algorithms for the diagnosis of CLD highlights the efficacy of various strategies. Notably, with consistently high accuracy rates, the RF model proved to be the most successful. Although preprocessing steps like replacing outliers and oversampling are important, the main focus of this study is the unified approach that combined non-linear (t-SNE, UMAP) and linear (LDA, FA) techniques for dimensionality reduction and feature integration. The reliability of our evaluations was guaranteed by carefully verifying and improving model performance using train-test splits and 10-fold cross-validation.

Subsequent studies in the field of chronic liver disease detection may explore sophisticated anomaly detection strategies and ensemble approaches, thereby expanding the potential to detect subtle indicators of the disease. The potential to strengthen the effectiveness of predictive models by adding extra clinical data beyond non-standard laboratory testing is promising. Furthermore, incorporating more dimensionality reduction techniques like PCA, and incremental feature selection could provide a more in-depth understanding of the features of the disease. Deep Belief Networks (DBNs), autoencoders, Non-negative Matrix Factorization (NMF), Random Forest using Particle Swarm Optimization (RF with PSO) could also be used. There is a chance that the detection of chronic liver disease will advance significantly with the use of these nuanced data analysis techniques, which seek to improve diagnostic precision and broaden the coverage of detection algorithms.

\begin{flushleft}
\textbf{Declaration}
\end{flushleft}

No subject testing or data collection procedures were taken into consideration for this study because it uses data that is publicly available or data from published sources. The authors declare unequivocally that all the work reported in this paper could not have been influenced by any known competing financial interest or personal relationship .\\\\
\textbf{Authors contribution}\\
All authors contributed equally.\\\\
\textbf{Availability of data and materials}\\
Data will be made available on request.

\section{References}
\begin{enumerate}
    \item Cheemerla, S., \& Balakrishnan, M. (2021b). Global Epidemiology of Chronic Liver Disease. Clinical Liver Disease, 17(5)
    \url{https://doi.org/10.1002/cld.1061}
    \item Asrani, S. K., Devarbhavi, H., Eaton, J. E., \& Kamath, P. S. (2019). Burden of liver diseases in the world. Journal of Hepatology, 70(1), 151–171. \url{https://doi.org/10.1016/j.jhep.2018.09.014}
     \item Umbare, R., Ashtekar, O., Nikhal, A., Pagar, B., \& Zare, O. (2023). Prediction \& Detection of Liver Diseases using Machine Learning. IEEE 3rd International Conference on TEMSMET.\\
     \url{https://doi.org/10.1109/temsmet56707.2023.10150135}
    \item Javaid, M., Haleem, A., Singh, R. P., Suman, R., \& Rab, S. (2022). Significance of machine learning in healthcare: Features, pillars and applications. International Journal of Intelligent Networks, 3, 58–73. \url{https://doi.org/10.1016/j.ijin.2022.05.002}
    \item Liu, Y., \& Chen, M. (2022). Epidemiology of liver cirrhosis and associated complications: Current knowledge and future directions. World Journal of Gastroenterology, 28(41), 5910–5930. \url{https://doi.org/10.3748/wjg.v28.i41.5910}
    \item Younossi, Z. M., Blissett, D. B., Blissett, R., Henry, L., Stepanova, M., Younossi, Y., Racila, A., Hunt, S., \& Beckerman, R. (2016). The economic and clinical burden of nonalcoholic fatty liver disease in the United States and Europe. Hepatology, 64(5), 1577–1586. \url{https://doi.org/10.1002/hep.28785}
    \item Popa, S., Ismaiel, A., Abenavoli, L., Padureanu, A. M., Dita, M. O., Bolchis, R., Munteanu, M., Brata, V. D., Pop, C., Bosneag, A., Dumitraşcu, D. I., Bârsan, M., \& David, L. (2023). Diagnosis of liver fibrosis using Artificial Intelligence: A Systematic review. Medicina-lithuania, 59(5), 992. \url{https://doi.org/10.3390/medicina59050992}
    \item Ahn, J., Connell, A., Simonetto, D. A., Hughes, C., \& Shah, V. H. (2021). Application of artificial intelligence for the diagnosis and treatment of liver diseases. Hepatology, 73(6), 2546–2563. \url{https://doi.org/10.1002/hep.31603}
    \item Bhupathi D, Tan C N-L, Tirumula S.S, Ray S.K. (2022) Liver Disease Detection using Machine Learning Techniques. The Computing and Information Technology Research and Education New Zealand (CITRENZ) Springer
    \item Mohan, V., Dhayanand, S. (2015). Liver Disease Prediction using SVM and Naïve Bayes Algorithms.
    \item Wu, J., Lin, S., Wan, B., Velani, B., \& Zhu, Y. (2019). Pyroptosis in Liver Disease: New Insights into Disease Mechanisms. Aging and Disease, 10(5), 1094. \url{https://doi.org/10.14336/ad.2019.0116}
    \item Amin, R., Yasmin, R., Ruhi, S., Rahman, M. H., \& Reza, M. S. (2023). Prediction of chronic liver disease patients using integrated projection based statistical feature extraction with machine learning algorithms. Informatics in Medicine Unlocked, 36, 101155. \url {https://doi.org/10.1016/j.imu.2022.101155}
    \item Batarseh, F. A., \& Freeman, L. (2022). AI Assurance: Towards Trustworthy, Explainable, Safe, and Ethical AI. Elsevier. ISBN 978-0-323-91919-7.
    \item Tavakol, M., \& Wetzel, A. P. (2020). Factor Analysis: a means for theory and instrument development in support of construct validity. International Journal of Medical Education, 11, 245–247. \url{https://doi.org/10.5116/ijme.5f96.0f4a}
    \item Cai, T., \& Ma, R. (2021). Theoretical Foundations of T-SNE for Visualizing High-Dimensional Clustered Data. arXiv (Cornell University). \url{https://doi.org/10.48550/arxiv.2105.07536}
    \item Singh, J., Bagga, S., \& Kaur, R. (2020). Software-based Prediction of Liver Disease with Feature Selection and Classification Techniques. Procedia Computer Science, 167, 1970–1980. \url{https://doi.org/10.1016/j.procs.2020.03.226}
    \item Wang, N., Yu, Y., Huang, D., Xu, B., Liu, J., Li, T., Xue, L., Zengyu, S., Chen, Y., \& Wang, J. (2015). Pulse Diagnosis signals analysis of fatty liver disease and cirrhosis patients by using machine learning. The Scientific World Journal, 2015, 1–9. \url{https://doi.org/10.1155/2015/859192}
    \item Kumar, Y., \& Sahoo, G. (2013). Prediction of different types of liver diseases using rule based classification model. Technology and Health Care, 21(5), 417–432. \url{https://doi.org/10.3233/thc-130742}
    \item Dritsas, I., \& Trigka, M. (2023). Supervised Machine learning models for liver disease risk prediction. Computers, 12(1), 19. \url{https://doi.org/10.3390/computers12010019}
    \item Muthuselvan, S., Rajapraksh, S., Somasundaram, K., \& Karthik, K. (2018). Classification of liver patient dataset using machine learning algorithms. International Journal of Engineering \& Technology, 7(3.34), 323. \url{https://doi.org/10.14419/ijet.v7i3.34.19217}
    \item Babu, M. S. P., Ramjee, M., Katta, S., \& Swapna, K. (2016). Implementation of partitional clustering on ILPD dataset to predict liver disorders. ICSESS. \url{https://doi.org/10.1109/icsess.2016.7883256}
    \item Sujith, J. G., Kumar, P., Reddy, S. J. M., \& Kanhe, A. (2023). Computative analysis of various techniques for classification of liver disease. Journal of Physics: Conference Series, 2466(1), 012035. \url{https://doi.org/10.1088/1742-6596/2466/1/012035}
    \item Sivasangari, A., Reddy, B. J. K., Kiran, A., \& Ajitha, P. (2020). Diagnosis of Liver Disease using Machine Learning Models. 2020 Fourth International Conference on I-SMAC (IoT in Social, Mobile, Analytics and Cloud) (I-SMAC). \url{https://doi.org/10.1109/i-smac49090.2020.9243375}
    \item McKnight, P. E., McKnight, K. M., Sidani, S., \& Figueredo, A. J. (2007). Missing data: A Gentle Introduction. Guilford Press
    \item Alpaydin, E. (2014). Introduction to machine learning. MIT Press.
    \item Muixí, A., Garcia-Gonzalez, A., Zlotnik, S., \& DiEz, P. (2023). Linear and nonlinear dimensionality reduction of biomechanical models. In Elsevier eBooks (pp. 23–44). \url{https://doi.org/10.1016/b978-0-32-389967-3.00004-4}
    \item Tharwat, A., Gaber, T., Ibrahim, A., \& Hassanien, A. E. (2017). Linear discriminant analysis: A detailed tutorial. Ai Communications, 30(2), 169–190. \url{https://doi.org/10.3233/aic-170729}
    \item Smelser, N. J., \& Baltes, P. B. (2001). International Encyclopedia of the Social \& Behavioral Sciences.
    \item Johnson, R. A., \& Wichern, D. W. (2007). Applied Multivariate Statistical analysis. Prentice Hall.
    \item Zhou, H., Wang, F., \& Tao, P. (2018). t-Distributed Stochastic Neighbor Embedding Method with the Least Information Loss for Macromolecular Simulations. Journal of Chemical Theory and Computation, 14(11), 5499–5510. \url{https://doi.org/10.1021/acs.jctc.8b00652}
    \item van der Maaten, L. J. P., \& Hinton, G. E. (2008). Visualizing High-Dimensional Data Using t-SNE. Journal of Machine Learning Research, 9(nov), 2579-2605.
    \item McInnes, L., Healy, J., \& Melville, J. (2018). UMAP: Uniform Manifold Approximation and Projection for dimensionality reduction. arXiv preprint arXiv:1802.03426.
    \item Vaidya, V., \& Vaidya, J. (2022). Impact of dimensionality reduction on Outlier Detection: an Empirical Study. IEEE 4th Int Conf Trust Priv Secur Intell Syst Appl (2022). \url{https://doi.org/10.1109/tps-isa56441.2022.00028}
    \item Halladin-Dabrowska, A., Kania, A., \& Kopeć, D. (2019). The T-SNE algorithm as a tool to improve the quality of reference data used in accurate mapping of heterogeneous Non-Forest vegetation. Remote Sensing, 12(1), 39. \url{https://doi.org/10.3390/rs12010039}
    \item Niha, S. A., Lakshmi, K. J., Blessi, P. G., Lakshmi, T. S., Chowdary, Y. M., \& Rao, M. P. (2023). A comparison of machine learning algorithms for predicting liver disease. International Journal for Research in Applied Science and Engineering Technology, 11(3), 2221–2230. \url{https://doi.org/10.22214/ijraset.2023.49954}
    \item Anthonysamy, V., \& Babu, S. K. K. (2023). Multi perceptron neural network and voting classifier for liver disease dataset. IEEE Access, 11, 102149–102156. \url{https://doi.org/10.1109/access.2023.3316515}
    \item Ghosh, M., Raihan, M. M. S., Raihan, M., Akter, L., Bairagi, A. K., Alshamrani, S. S., \& Masud, M. (2021). A comparative analysis of machine learning algorithms to predict liver disease. Intelligent Automation and Soft Computing, 30(3), 917–928. \url{https://doi.org/10.32604/iasc.2021.017989}
    \item Panwar, V., Choudhary, N., Mittal, S., \& Sahu, G. (2021). Review of liver disease prediction using machine learning algorithm. Journal of Emerging Technology and Innovative Research (JETIR), 8(2), 2349–5162.
    \item Kuzhippallil, M. A., Joseph, C., \& Kannan, A. (2020). Comparative Analysis of Machine Learning Techniques for Indian Liver Disease Patients. ICACCS. \url{https://doi.org/10.1109/icaccs48705.2020.9074368}
\end{enumerate}


\printbibliography 


\end{document}